\documentclass{article} 
\usepackage[preprint]{colm2026_conference}
\usepackage{microtype}
\usepackage{hyperref}
\usepackage{url}
\usepackage{booktabs}

\usepackage{wrapfig}
\usepackage{pdfpages}
\usepackage{amsmath}
\usepackage{graphicx}
\usepackage{multirow}
\usepackage{caption}
\usepackage{enumitem}
\usepackage{subcaption}
\usepackage[utf8]{inputenc}
\usepackage[T1]{fontenc}
\usepackage{silence}
\usepackage[pass]{geometry}
  \usepackage[table]{xcolor}
  \definecolor{darkgreen}{rgb}{0.0, 0.55, 0.25}
  \definecolor{deltared}{rgb}{0.75, 0.0, 0.0}
  \definecolor{rowhl}{RGB}{232, 241, 255}  
  \newcommand{\gain}[1]{\hspace{4pt}{\color{darkgreen}{\scriptsize $\uparrow$}\hspace{2pt}{\scriptsize #1}}}
  \newcommand{\loss}[1]{\hspace{4pt}{\color{deltared}{\scriptsize $\downarrow$}\hspace{2pt}{\scriptsize #1}}}

\usepackage[many]{tcolorbox} 
\tcbuselibrary{listings}

\usepackage{pdfrender}
\newcommand*{\superbold}[1]{%
  \textpdfrender{%
    TextRenderingMode=FillStroke,
    LineWidth=0.66pt,
  }{#1}%
}

\usepackage{lineno}

\definecolor{darkblue}{rgb}{0, 0, 0.5}
\hypersetup{colorlinks=true, citecolor=darkblue, linkcolor=darkblue, urlcolor=darkblue}

\title{Entropy Centroids as Intrinsic Rewards for Test-Time Scaling}

\author{
    Wenshuo Zhao$^{1,2}$
    Qi Zhu$^{3}$
    Xingshan Zeng$^{3}$
    Fei Mi$^{3}$
    Lifeng Shang$^{3}$ 
    Yi R. (May) Fung$^{1}$\\
    \\
    $^{1}$HKUST \quad
    $^{2}$ZJU \quad
    $^{3}$Huawei \\
    \texttt{wzhaoba@connect.ust.hk}
  }

\begin{document}

\ifcolmsubmission
\linenumbers
\fi

\maketitle

\begin{abstract}

An effective way to scale up test-time compute of large language models is to sample multiple responses and then select the best one, as in Grok Heavy and Gemini Deep Think.  
Existing selection methods often rely on external reward models, which requires training a strong reward model and introduces additional computation overhead. As an alternative, previous approaches have  explored intrinsic signals, such as confidence and entropy, but these signals are noisy with naive aggregation. In this work, we observe that high-entropy tokens tend to cluster into consecutive groups during inference, providing a more stable notion of model uncertainty than individual tokens. Together, these clusters reveal temporal patterns of model uncertainty throughout the inference process. Motivated by this observation, we propose to use the temporal structure of uncertainty as an intrinsic reward. To this end, we first formalize the basic unit of segment-level uncertainty as the High Entropy Phase (HEP), a variable-length segment that begins at a high-entropy token and ends when  consecutive low-entropy tokens appear. We then define the Entropy Centroid, inspired by the concept of the center of mass in physics, as the weighted average position of all HEPs along the trajectory. Intuitively, a lower centroid indicates early exploration followed by confident generation, which we find often corresponds to higher response quality. Based on this insight, we propose the Lowest Centroid method, which selects the response with the lowest entropy centroid among multiple candidates. Experiments on mathematics, code generation, logical reasoning, and agentic tasks, across model scales ranging from 14B to 480B, show that Lowest Centroid consistently outperforms existing baselines and delivers stable gains as model size increases. Code is available at \url{https://github.com/hkust-nlp/entropy-centroid}.

\end{abstract}

\section{\textsc{Introduction}} \label{intro}

\begin{figure}[!t] 
  \centering 
  \includegraphics[width=\textwidth]{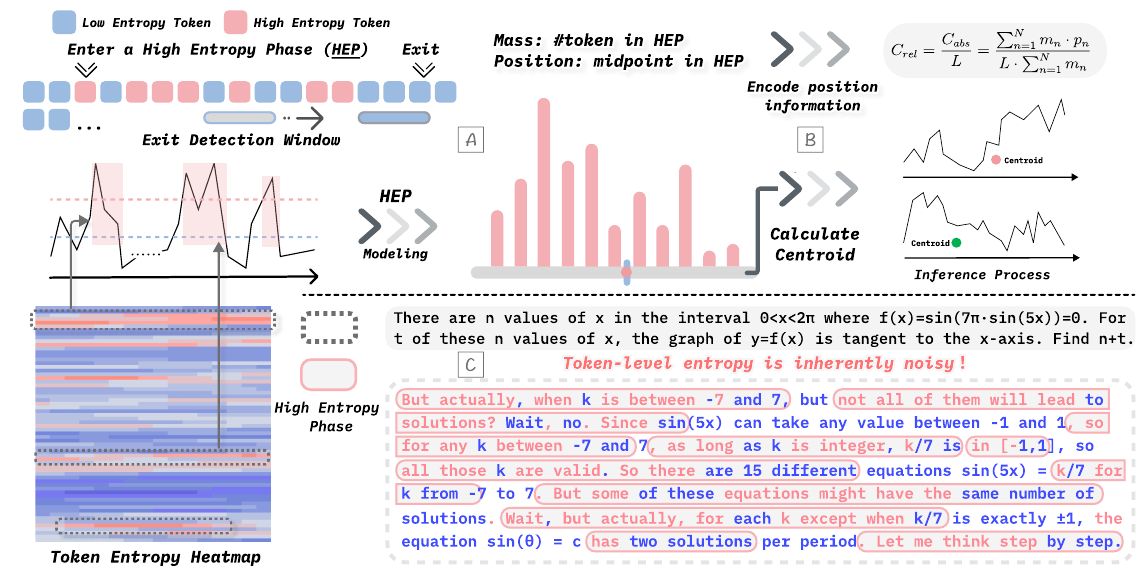} 
  \caption{Overview of the Entropy Centroid framework. {\bf Top panel:} a) High Entropy Phase (HEP). When a token with entropy above the top entropy threshold is detected, the model enters a High Entropy Phase. The phase ends once $k$ consecutive low-entropy tokens are observed. b) Entropy Centroid. For each HEP, we treat the number of tokens as its mass and the midpoint as its position, then compute the weighted average position along the trajectory. The formula follows the center-of-mass in physics. {\bf Bottom panel:} c) We visualize the token-level entropy evolution. Scattered high-confidence tokens frequently appear within critical decisions, making token-level entropy a noisy reward signal. HEP (each red box highlights one phase) filters out such single-token noise at a higher group level.}
  \label{fig:teaser}
\end{figure}

Reward signals play a central role in test-time scaling of large language models (LLMs), where in parallel scaling, $N$ candidate trajectories are sampled and the best one is selected (Gemini Deep Think~\citep{feng2026towards}, Grok 4 Heavy~\citep{grok4heavy2026}). Existing approaches typically rely on an external reward model to score each trajectory or on expensive human annotation~\citep{lightman2023letsverifystepstep,wang2024mathshepherdverifyreinforcellms}, which introduce substantial computational overhead~\citep{fei2025self,chen2025spectral,cao2025more} and noisy step-level evaluation~\citep{ding2025scan,zhang2025groundedprm,xie2026towards}. These costs motivate the search for intrinsic reward signals derived from the model's own generation process~\citep{yuan2024self,zelikman2024quiet}. Majority voting~\citep{wang2022selfconsistency} is a simple example, but it is restricted to short-answer tasks where the outputs can be easily aggregated, and does not generalize to complex settings such as coding and agentic tasks.

Recent work moves beyond voting and examines token-level metrics — entropy, confidence, and KL divergence — as proxies for inference quality~\citep{kang2025scalable,zhao2025learning}. However, the token-level signals are inherently noisy. As shown in Figure~\ref{fig:teaser} (c), calculation tokens tend to be confident even when the model is actually uncertain about the step overall. Several methods~\citep{fu2025deep,sharma2025think} address this by aggregating token-level signals over fixed-length local windows, such as averaging the confidence scores over 1,024 tokens as an indicator of trajectory quality. However, these methods apply a monotonic rule --- low entropy is uniformly treated as being confident and high entropy as failure, regardless of position in the trajectory. This view is challenged by recent findings that high-entropy tokens can reflect productive exploration rather than errors~\citep{wang2025beyond,hu2026entropy}.

Hypothesizing that the temporal pattern of uncertainty over a trajectory may contain useful information, in this work, we explore temporal uncertainty as an intrinsic reward.

A natural choice is to model token-level entropy changes over a trajectory. However, as shown in Figure~\ref{fig:teaser}(c), token-level entropy tends to be noisy, which may misrepresent the model's certainty within the current reasoning unit. As such, we seek a coarser-grained unit to represent the model's uncertainty state.
A closer look at token-level entropy reveals a simple but informative structure: As the heatmaps in Figure~\ref{fig:teaser} show, high-entropy tokens do not appear in isolation; they cluster into consecutive bursts. When a model encounters difficulties or decision points, it enters a period of high entropy that resolves either into confident low-entropy continuation or into persistent uncertainty. Therefore, we formalize this pattern with \textbf{High Entropy Phases (HEPs)} as a basic unit to represent the model's uncertainty state that could happen in different positions in the trajectory. Specifically, HEP is a variable-length unit, which begins when the token entropy exceeds a high threshold and ends when a specified number of consecutive tokens fall below a low threshold.

\begin{figure*}[!t]
  \centering
  \begin{subfigure}[b]{0.68\textwidth}
    \centering
    \includegraphics[width=\textwidth]{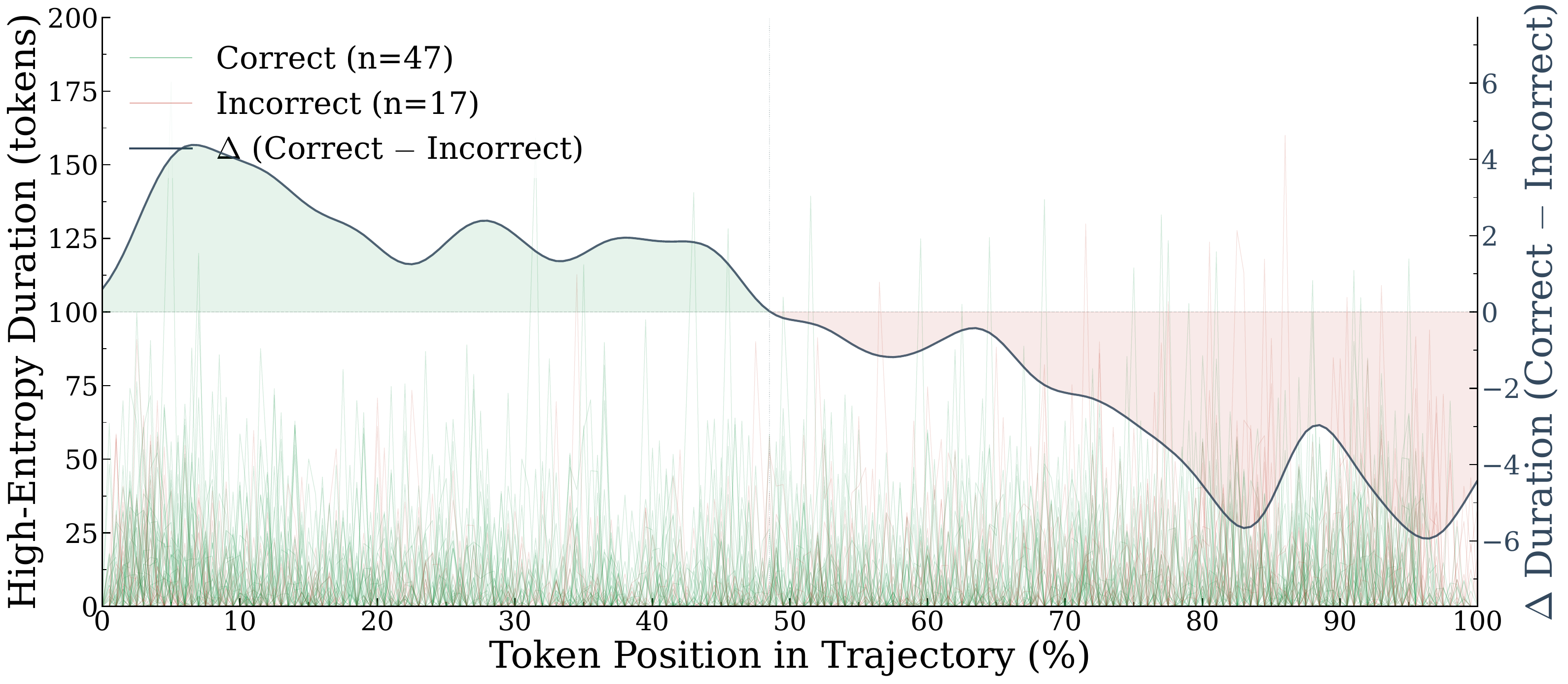}
    \label{fig:hep}
  \end{subfigure}
  \hfill
  \begin{subfigure}[b]{0.3\textwidth}
    \centering
    \includegraphics[width=0.9\textwidth]{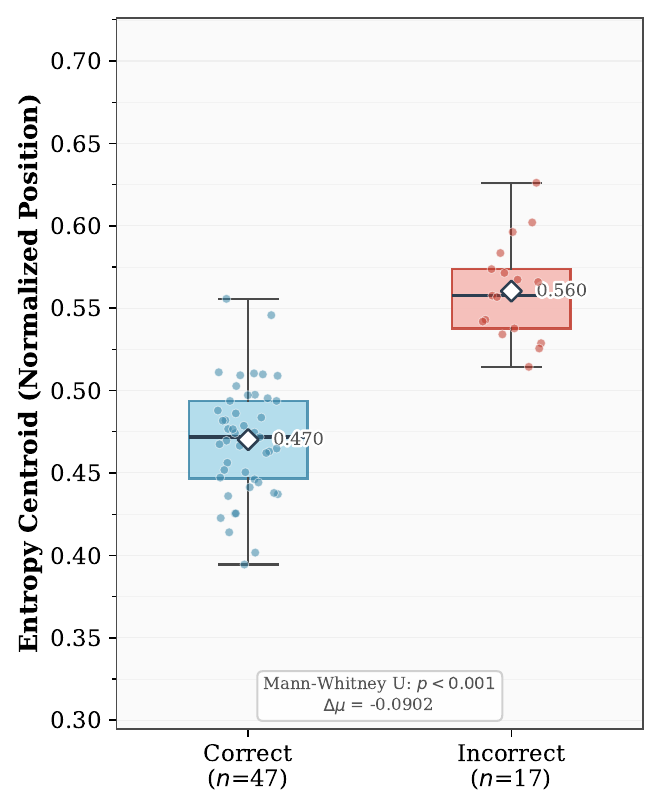}
    \label{fig:centroid}
  \end{subfigure}
  \caption{\textbf{Left:} We sample 64 responses for the same question and
  visualize the model's inference over time using HEP. The duration on the Y-axis refers to the number of tokens in each HEP. The green curves represent correct trajectories and the red curves represent incorrect ones. We compute the difference between the mean curves of the two groups and apply Gaussian smoothing to reveal the overall trend. The results show a clear separation: correct trajectories concentrate more HEPs in the early stage, while incorrect trajectories accumulate HEPs in the later stage, gradually surpassing the correct ones as inference progresses. \textbf{Right:} Based on HEP modeling, we compute the entropy centroids. Correct trajectories (blue scatter) have a median centroid around 0.47, while incorrect trajectories (red scatter) have a median centroid around 0.55.}
  \label{fig:statistic}
\end{figure*}

With HEPs as basic units, we further encode their positional information along the trajectory. Borrowing the concept of center of mass from physics, we define the \textbf{Entropy Centroid} by analogy. For each HEP, we take the number of tokens as its mass and the midpoint as its position, then compute the weighted average position along the trajectory (formulated in Figure~\ref{fig:teaser} (b)). We validate this design empirically. Figure~\ref{fig:statistic} visualizes HEP-level state transitions across trajectories. A clear separation emerges: correct trajectories tend to exhibit more exploration in the early stages, while incorrect trajectories show more confusion in the later stages. This is consistent with our goal of capturing temporal patterns in inference. We further plot the centroid distributions for different trajectories on the same problem. Figure~\ref{fig:statistic} confirms that centroid values of correct and incorrect trajectories are well separated, with correct trajectories concentrating exploration earlier. This supports the use of the lowest centroid as a selection criterion.

We use the Entropy Centroid as the intrinsic reward in test-time scaling. Given $N$ sampled trajectories for a problem, we simply select the one with the lowest Entropy Centroid. We evaluate the Lowest Centroid selection strategy across model scales from 14B to 480B parameters on mathematics, code generation, logical reasoning, and agentic tasks. Across all settings, Lowest Centroid consistently outperforms existing baselines (Section~\ref{experiments}), with gains that remain stable as model size increases, demonstrating that the Entropy Centroid is a general-purpose intrinsic reward signal for test-time scaling.

\section{\textsc{Basic Units for Measuring Inference Quality}} \label{sec:hep}

Our motivation is to capture temporal patterns along the trajectory. Before defining a method for encoding temporal information, we first need to propose a suitable basic unit to characterize inference state.

Token-level entropy is an intuitive measure of model uncertainty, but it is unreliable as a reward signal~\citep{gupta2024language} for two reasons. First, as discussed in Section~\ref{intro}, the model inherently assigns high confidence to tokens involving numerical calculations or simple sentence completions, and the trajectory-level average score loses the temporal structure. Second, supervised fine-tuning and reinforcement learning tend to concentrate prediction probability on high-reward token sequences~\citep{lu2025onpolicydistillation}. This mode-covering behavior causes token-level confidence to reflect training incentives rather than true inference quality~\citep{gao2023scaling,slocum2025diverse}, partially distorting the output logit distributions. To obtain a stable signal, we must elevate the measurement unit from single tokens to token groups.

\textbf{Definition of High Entropy Phase (HEP).} Let $\mathcal{T} = \{t_1, t_2, \dots, t_L\}$ denote the token entropy sequence of length $L$ generated during model inference. We define $\theta_{high}$ as the top entropy threshold and $\theta_{low}$ as the bottom entropy threshold over the sequence. To formalize the boundaries of a High Entropy Phase, we introduce a state variable $S_i \in \{0, 1\}$ to denote the phase state at the $i$-th token, where $S_i = 1$ indicates that the model is currently in a HEP. The state transition is defined recursively as follows:
\begin{equation} 
S_i = \begin{cases} 
1, & \text{if } S_{i-1} = 0 \text{ and } t_i \ge \theta_{high} \\ 
0, & \text{if } S_{i-1} = 1 \text{ and } \sum_{j=0}^{k-1} \mathbb{I}(t_{i-j} \le \theta_{low}) = k \\ 
S_{i-1}, & \text{otherwise} 
\end{cases} 
\end{equation}
 Here, $\mathbb{I}(\cdot)$ denotes the indicator function. The model begins in the inactive state $S_0 = 0$, i.e., outside any HEP. A HEP is triggered whenever the model is inactive ($S_{i-1} = 0$) and encounters a token whose entropy reaches the threshold ($t_i \ge \theta_{high}$), upon which the state transitions to $S_i = 1$. To determine phase exit, we employ a sliding window of size $k$: a HEP terminates ($S_i$ resets to $0$) only when the most recent $k$ consecutive tokens all fall below the lower threshold $\theta_{low}$. In all other cases, the current state persists. This recursive procedure applies identically throughout the remainder of the sequence.

 A contiguous sub-sequence of tokens where $S_i = 1$ constitutes a single HEP. Unlike fixed-length windows that impose arbitrary boundaries, A HEP provide a natural and adaptive unit for representing the model's inference state at each stage of inference, preserving the integrity of each uncertain phase as a coherent unit.

\section{\textsc{Entropy Centroid}} \label{sec:method}

The High Entropy Phase (HEP) formulation captures the extent of uncertainty at each inference step via phase mass, but a key question remains: \textit{how should we encode the temporal information of these phases to assess overall inference quality?}

\subsection{Entropy-Based Centroid Computation}
\label{sec:centroid-compute}

Inspired by the center of mass in physics, we define the Entropy Centroid to summarize where uncertainty concentrates across the generation sequence. Each HEP serves as the basic unit in this computation. For the $n$-th HEP in the sequence, denoting its start index as $a_n$ and end index as $b_n$, its mass $m_n$ and position $p_n$ are formulated as follows:

\begin{equation} 
m_n = b_n - a_n + 1, \quad p_n = \frac{a_n + b_n}{2} 
\end{equation}

That is, $m_n$ measures the length (in tokens) of the $n$-th HEP, while $p_n$ locates its midpoint within the sequence. Note that here we treat every token within HEPs equally with a mass of one, and do not use their specific entropy values. As discussed in Section~\ref{sec:hep}, token-level entropy values are inherently noisy, which motivates us to adopt this coarse-grained strategy.

Assuming $N$ HEPs are detected in the full sequence, 
the position (normalized by the total sequence length $L$) of the centroid is defined as:

\begin{equation}
\label{eq:centroid}
\quad C = \frac{\sum_{n=1}^{N} m_n \cdot p_n}{L \cdot \sum_{n=1}^{N} m_n}
\end{equation}

Trajectory lengths vary across samples, so we normalize the centroid by the total sequence length $L$ to obtain a scale-invariant measure $C \in (0, 1)$, ensuring that centroid values remain comparable regardless of trajectory length. A lower $C$ indicates that the uncertainty of the model concentrates in the early stages of inference, while a higher $C$ indicates uncertainty concentrated toward the later stages. Since $C$ is the primary quantity used throughout our analysis, we refer to it simply as the \emph{centroid} hereafter.

While it seems we ignore tokens outside HEPs when computing the entropy centroid, our definition in Eq.~\ref{eq:centroid} essentially models entropy of all tokens as the mass to compute the centroid, but adopts a binary mass approximation --- we approximate the mass of those low-entropy tokens outside HEPs as zero, and the mass of each token within HEPs as one. This is to avoid potential noises from the specific entropy values. In Section \ref{sec:abla}, we will compare with entropy centroid computed from raw entropy values empirically.

\subsection{Lowest-Centroid Selection for Test-Time Scaling}

The centroid calculated from the High Entropy Phases reflects the certainty of the model during the inference process. We plot boxplots in Figure~\ref{fig:statistic} to illustrate the centroid distribution across different trajectories. The figures clearly demonstrate significant differences in the centroid distributions between correct and incorrect trajectories. Trajectories with correct answers typically exhibit an inference pattern of exploration followed by exploitation. Incorrect trajectories show excessive early exploitation, which leads the subsequent inference into confusion.

 Entropy Centroid reflects the exploration-exploitation trade-off in sequential generation. Early in the inference trace, the model must explore diverse inference paths. High entropy at this stage reflects productive exploration. As generation progresses toward a conclusion, the model should exploit its accumulated inference and converge. High entropy at this stage instead signals unresolved confusion. A trajectory whose uncertainty concentrates early is therefore more likely to yield a correct answer than one whose uncertainty persists late.

Therefore, we propose Lowest Centroid based on the Entropy Centroid as a trajectory selection strategy in test-time scaling. This approach characterizes the intrinsic features of the model during inference. The implementation of this method is straightforward and highly effective.

\textbf{Our method straightforwardly selects the trajectory with the lowest centroid from multiple candidate trajectories.} Issues such as abnormal termination, repetitive outputs, and gibberish generation may occur during inference. The entropy distributions of these specific trajectories differ significantly from those of normal inference trajectories. Consequently, we first perform an outlier filtering step to remove trajectories that deviate significantly from the mean centroid of all candidates, and then we select the trajectory with the lowest centroid from the remaining set.

\section{\textsc{Experiments}} \label{experiments}

\begin{table}[!t]
    \centering
    \renewcommand{\arraystretch}{1.2} 
    \setlength{\tabcolsep}{4pt}
    \resizebox{\textwidth}{!}{
 \begin{tabular}{cccccccc}
          \toprule[1.2pt]
           \textbf{Acc $(\%,\uparrow)$} & \textbf{Method} & \multicolumn{1}{c}{\textbf{Agent@64}} & \multicolumn{2}{c}{\textbf{Code@32}} & \multicolumn{2}{c}{\textbf{Math@64}} &
  \multicolumn{1}{c}{\textbf{Logic@64}} \\

          \cmidrule(lr){3-3} \cmidrule(lr){4-5} \cmidrule(lr){6-7} \cmidrule(lr){8-8}

       \textbf{Model} &  \textbf{Dataset} & \ \textbf{$\tau^2$-Bench}  & \ \textbf{BigCode} & \ \textbf{LiveCode} & \ \textbf{AIME25} & \ \textbf{Minerva} &
  \ \textbf{Synlogic} \\

          \midrule[0.8pt]

          \multirow{6}{*}{\shortstack[c]{Olmo-3.1\\-Think}}
          & Pass@1 & 9.5  & 42.6 & 74.2 & 75.7 & 55.0 & 42.1  \\
          & {\footnotesize Greedy Decoding} & 7.0\loss{2.5}  & \superbold{44.9}\gain{2.3} & 71.6\loss{2.6} & 70.0\loss{5.7}  & 54.8\loss{0.2} & 40.3\loss{1.8}  \\
          & {\footnotesize Self-Certainty}  & 7.0\loss{2.5}  & 43.3\gain{0.7} & 76.6\gain{2.4} & 76.7\gain{1.0}  & 54.4\loss{0.6} & 42.3\gain{0.2}  \\
          & {\footnotesize Tail Confidence}  & 4.0\loss{5.5}  & 41.5\loss{1.1} & 76.3\gain{2.1} & 73.3\loss{2.4}  & 54.4\loss{0.6} & 46.0\gain{3.9} \\
          & {\footnotesize Bottom Window}  & 8.0\loss{1.5}  & 43.4\gain{0.8} & 75.9\gain{1.7} & 80\gain{4.3}  & 54.8\loss{0.2} & 48.9\gain{6.8} \\
          \rowcolor{rowhl}
          & {\footnotesize Lowest Centroid} & \superbold{12.0}\gain{2.5}  & 44.5\gain{1.9} & \superbold{77.2}\gain{3.0} & \superbold{86.7}\gain{11.0}  & \superbold{55.9}\gain{0.9} & \superbold{50.0}\gain{7.9} \\

          \midrule
          \multirow{6}{*}{\shortstack[c]{Qwen3\\-14B}}
          & Pass@1 & 25.9  & 39.9 & 77.2 & 71.4  & 56.3 & 44.4 \\
          & {\footnotesize Greedy Decoding} & 29.0\gain{3.1} & 39.8\loss{0.1} & 72.6\loss{4.6} & 70.0\loss{1.4}  & 55.5\loss{0.8} & 38.9\loss{5.5}  \\
          & {\footnotesize Self-Certainty}  & 18.0\loss{7.9}  & \superbold{46.2}\gain{6.3} & \superbold{79.5}\gain{2.3} & 73.3\gain{1.9} & 55.2\loss{1.1} & \superbold{50.3}\gain{5.9} \\
          & {\footnotesize Tail Confidence}  & 21.0\loss{4.9}  & 45.4\gain{5.5} & 78.6\gain{1.4} & 70.0\loss{1.4}  & 55.5\loss{0.8} & \superbold{50.3}\gain{5.9} \\
          & {\footnotesize Bottom Window}  & 23.0\loss{2.9}  & 44.0\gain{4.1} & 77.7\gain{0.5} & 73.3\gain{1.9}  & 55.9\loss{0.4} & 48.9\gain{4.5} \\
          \rowcolor{rowhl}
          & {\footnotesize Lowest Centroid} & \superbold{35.0}\gain{9.1}  & 46.0\gain{6.1} & 78.1\gain{0.9} & \superbold{76.7}\gain{5.3}  & \superbold{58.1}\gain{1.8} & 45.4\gain{1.0} \\

          \midrule
          \multirow{6}{*}{QWQ-32B}
          & Pass@1 & 12.2  & 43.9 & 76.7 & 68.5  & 54.5 & 46.0 \\
          & {\footnotesize Greedy Decoding} & 12.0\loss{0.2}  & 43.4\loss{0.5} & 73.2\loss{3.5} & 56.7\loss{11.8}  & 53.3\loss{1.2} & 40.9\loss{5.1} \\
          & {\footnotesize Self-Certainty}  & 9.0\loss{3.2}  & 44.0\gain{0.1} & \superbold{78.1}\gain{1.4} & 70.0\gain{1.5}  & 54.0\loss{0.5} & 48.6\gain{2.6} \\
          & {\footnotesize Tail Confidence}  & 8.0\loss{4.2} & 42.3\loss{1.6} & 77.4\gain{0.7} & 76.7\gain{8.2}  & 54.8\gain{0.3} & 44.6\loss{1.4} \\
          & {\footnotesize Bottom Window}  & 9.0\loss{3.2}  & 44.1\gain{0.2} & 77.9\gain{1.2} & 70.0\gain{1.5} & 55.9\gain{1.4} & 55.7\gain{9.7}  \\
          \rowcolor{rowhl}
          & {\footnotesize Lowest Centroid} & \superbold{18.0}\gain{5.8}  & \superbold{44.5}\gain{0.6} & 78.0\gain{1.3} & \superbold{83.3}\gain{14.8}  & \superbold{57.0}\gain{2.5} & \superbold{56.0}\gain{10.0} \\

          \midrule
          \multirow{6}{*}{\shortstack[c]{Ministral\\-3-14B\\-Instruct}}
          & Pass@1 & 26.5  & 22.9 & 40.3 & 27.5  & 46.7 & 23.1  \\
          & {\footnotesize Greedy Decoding} & \superbold{33.0}\gain{6.5}  & \superbold{26.3}\gain{3.4} & 39.7\loss{0.6} & 30.0\gain{2.5} & 48.2\gain{1.5} & 16.3\loss{6.8}  \\
          & {\footnotesize Self-Certainty}  & 22.0\loss{4.5} & 18.0\loss{4.9} & 40.0\loss{0.3} & 43.3\gain{15.8} & 48.2\gain{1.5} & 20.0\loss{3.1}  \\
          & {\footnotesize Tail Confidence}  & 21.0\loss{5.5} & 19.1\loss{3.8} & \superbold{46.8}\gain{6.5} & \superbold{46.7}\gain{19.2}  & 46.0\loss{0.7} & \superbold{38.9}\gain{15.8}  \\
          & {\footnotesize Bottom Window}  & 21.0\loss{5.5}  & 19.0\loss{3.9} & 44.5\gain{4.2} & 36.7\gain{9.2}  & 46.3\loss{0.4} & 36.6\gain{13.5}  \\
          \rowcolor{rowhl}
          & {\footnotesize Lowest Centroid} & \superbold{33.0}\gain{6.5}  & 25.5\gain{2.6} & 44.4\gain{4.1} & 40.0\gain{12.5}  & \superbold{50.4}\gain{3.7} & 33.4\gain{10.3} \\

          \midrule
          \multirow{6}{*}{\shortstack[c]{GPT-OSS\\-120B}}
          & Pass@1 & 54.9  & 35.0 & 81.2 & 76.5  & 38.6 & 63.1  \\
          & {\footnotesize Greedy Decoding} & 58.0\gain{3.1}  & 36.1\gain{1.1} & 82.0\gain{0.8} & 76.7\gain{0.2} & 37.9\loss{0.7} & 62.0\loss{1.1} \\
          & {\footnotesize Self-Certainty}  & 57.0\gain{2.1}  & 34.9\loss{0.1} & 80.3\loss{0.9} & 70.0\loss{6.5}  & 36.0\loss{2.6} & 61.4\loss{1.7} \\
          & {\footnotesize Tail Confidence}  & 53.0\loss{1.9} & \superbold{36.8}\gain{1.8} & 80.1\loss{1.1} & 70.0\loss{6.5}  & 38.2\loss{0.4} & 66.0\gain{2.9} \\
          & {\footnotesize Bottom Window}  & 49.0\loss{5.9} & 36.1\gain{1.1} & 81.3\gain{0.1} & 73.3\loss{3.2}  & 38.2\loss{0.4} & 64.3\gain{1.2}  \\
          \rowcolor{rowhl}
          & {\footnotesize Lowest Centroid} & \superbold{65.0}\gain{10.1}  & 35.8\gain{0.8} & \superbold{83.6}\gain{2.4} & \superbold{90.0}\gain{13.5}  & \superbold{40.1}\gain{1.5} & \superbold{67.4}\gain{4.3}  \\

          \bottomrule[1.2pt]
      \end{tabular}}
      \caption{Benchmarking Results across Agent, Code, Math, and Logic Tasks. All experiments use a unified set of HEP parameters. For each model–dataset pair, we generate a shared evaluation cache. All subsequent trajectory selections are drawn from this cache, ensuring a fair and consistent comparison across methods. We annotate the absolute change compared to the Pass@1 baseline in the table. Our method consistently outperforms the Pass@1 baseline across all settings.}
    \label{tab:main_results}
\end{table}

\subsection{Experimental Setup}

\paragraph{Models.} 
To validate the generality of the entropy centroid method across various LLM architectures, we evaluate a wide array of model families with parameter scales ranging from 14B to 480B. This selection includes long chain-of-thought (CoT) reasoning models and advanced models specialized in programming and agentic workflows, specifically Olmo-3.1-32B-Thinking~\citep{olmo2025olmo3}, GPT-OSS-120B~\citep{openai2025gptoss120bgptoss20bmodel}, Qwen3-14B~\citep{qwen3technicalreport}, QwQ-32B~\citep{qwq32b}, Ministral-3-14B~\citep{liu2026ministral3}, Qwen3-Coder-480B-A35B-Instruct-FP8~\citep{qwen3technicalreport}, Minimax M2.5~\citep{chen2025minimax} and Qwen3-Coder-Next~\citep{qwen_qwen3_coder_next_tech_report}. More experiment details are in Appendix~\ref{app:exp}.

\paragraph{Benchmarks.} The evaluation involves comprehensive benchmarks across diverse domains: (1) \textbf{Math}. We use AIME 2025~\citep{aime25} and Minerva Math~\citep{lewkowycz2022solving}, which represent challenging mathematics competition-level datasets. (2) \textbf{Logic}. We use Synlogic \citep{liu2025synlogic} to assess complex logical tasks, which include sudoku, arrow maze and puzzles. (3) \textbf{Code.} Code generation performance, including self-repair and execution, is measured on competitive programming tasks from BigCodeBench \citep{zhuo2024bigcodebench} and LiveCodeBench \citep{jain2024livecodebench}. (4) \textbf{Agent.} We investigate agentic capabilities using $\tau^2$-Bench~\citep{barres2025tau}, requiring models to perform multi-step reasoning, planning, and tool use to reach specific goals without human intervention. 

\paragraph{Setup.}
We evaluate our approach on datasets spanning multiple domains, including agentic tasks, coding, mathematics, and logical reasoning. To ensure experimental rigor, we generate an evaluation cache after model inference. All subsequent methods read from the same cache, guaranteeing evaluation consistency across different selection strategies. We sample 64 responses for the mathematics, logic, and agent tasks, and 32 trajectories for the coding tasks to keep computational overhead manageable. To define the high entropy phase, we set the top entropy token threshold $\theta_{high}$ to $1\%$ and the bottom entropy token threshold $\theta_{low}$ to $80\%$. The number of consecutive tokens required to exit the high entropy phase is set to $k=2$. These hyperparameters regarding HEP computation are kept the same for all models across all benchmarks in our experiments. When computing the entropy of each token in the sequence, we only use the top 10 tokens in the vocabulary with the highest probability, as they already capture most of the probability mass in the output distribution.

\paragraph{Baselines.}
We report Pass@1 and Greedy Decoding~\citep{holtzman2019curious} as a reference for model capability. In addition, we compare against several intrinsic-reward methods proposed for test-time scaling: (1) \textbf{Self-Certainty}~\citep{kang2025scalable} measures the KL divergence between the model's output distribution and the uniform distribution, using the resulting score to select among candidate trajectories. (2) \textbf{Tail Confidence and Bottom Window}~\citep{fu2025deep} aggregate token-level confidence scores over contiguous groups of tokens (e.g., the lowest-confidence window) and originally serve as weights for majority voting. Because majority voting is restricted to short-answer tasks and thus incompatible with our evaluation setting, we repurpose these scores as trajectory-selection criteria instead.

\subsection{Lowest Entropy Centroid Evaluations}

\paragraph{Main Results.}
Table~\ref{tab:main_results} presents the main experimental results. Entropy centroid outperforms other intrinsic reward methods in most cases, achieving an average improvement of 5.3\% over Pass@1 across all 30 model-dataset combinations. On agentic tasks, the average improvement reaches 6.8\%, where GPT-OSS-120B achieves a 10.1\% gain over Pass@1. Our method is the only one that outperforms Pass@1 across all settings. In contrast, existing baselines exhibit unstable behavior. For example, Self-Certainty outperforms Pass@1 on only two out of six datasets for Ministral-3-14B-Instruct. Tail Confidence and Bottom Window perform poorly on agentic tasks, where all results fall below the Pass@1 baseline. These comparisons highlight a key advantage of our approach: Rather than excelling on specific settings while failing on others, Lowest Centroid delivers reliable gains across diverse domains and model scales.

\begin{wraptable}{r}{0.45\textwidth}
\centering
\vspace{-10pt}
    \renewcommand{\arraystretch}{1.2} 
    \setlength{\tabcolsep}{3pt} 
     \resizebox{0.45\textwidth}{!}{\begin{tabular}{l @{\hspace{1.5pt}} *{3}{c}}
      \toprule[1.2pt]

      \multirow{2}{*}{\textbf{Method}}
      & \scriptsize \textbf{Qwen3-Coder}
      & \scriptsize \textbf{Qwen3-Next}
      & \scriptsize \textbf{Minimax-M2.5} \\

      & \scriptsize \textbf{480BA35B}
      & \scriptsize \textbf{80BA3B}
      & \scriptsize \textbf{230BA10B} \\
      \midrule[0.8pt]

      Pass@1    & 58.5 & 20.1 & 66.1  \\
      {\footnotesize Greedy Decoding} & 56.0\loss{2.5} & 27.0\gain{6.9} & 65.0\loss{1.1} \\
      {\footnotesize Tail Confidence} & 50.0\loss{8.5} & 2.0\loss{18.1} & 32.0\loss{33.9}  \\
      {\footnotesize Bottom Window}  & 53.0\loss{5.5} & 9.0\loss{11.1} & 48.0\loss{18.1} \\
      \rowcolor{rowhl}
      {\footnotesize Lowest Centroid} & \superbold{65.0}\gain{6.5} & \superbold{27.0}\gain{6.9} & \superbold{73.0}\gain{6.9} \\

      \bottomrule[1.2pt]
  \end{tabular}}
    \captionof{table}{$\tau^2$-bench results for large models.}
    \label{tab:tau2_big_model}
    \vspace{-10pt}
\end{wraptable}

\paragraph{Results on Large Agents.} 
Most models in our main experiments are below 50B parameters. Since smaller models perform poorly on agentic tasks and fail to produce meaningful performance differentiation, we additionally evaluate three larger models on this domain. Agentic tasks represent one of the most actively studied areas in current research, making this a valuable testbed for generalization. As shown in Table~\ref{tab:tau2_big_model}, the Entropy Centroid continues to deliver strong performance at this scale.

\paragraph{Comparison with Majority Voting.}
While we believe majority voting is not a practical method as it is only applicable to short-answer tasks, we still compare Lowest Centroid against majority voting on tasks where extracted answers are available. As shown in Table~\ref{tab:majority_voting} of Appendix~\ref{app:exp}, the two methods achieve comparable overall performance across different datasets and models. Lowest Centroid has a notable advantage on Olmo-3.1-32B-Thinking (80.0\%$\rightarrow$ 86.7\%, 46.9\%$\rightarrow$ 50.0\%) and QWQ-32B (80.0\%$\rightarrow$ 83.3\%, 48.0\%$\rightarrow$ 56.0\%) for AIME25 and Synlogic. 
This result is significant because majority voting relies on answer aggregation, which is only applicable to tasks with short, verifiable answers.
We would like to emphasize our approach has much broader usage as already shown in agentic and coding tasks.

\subsection{Ablation Study}\label{sec:abla}

\begin{figure}[!t]
  \centering
    \includegraphics[width=0.5\textwidth]{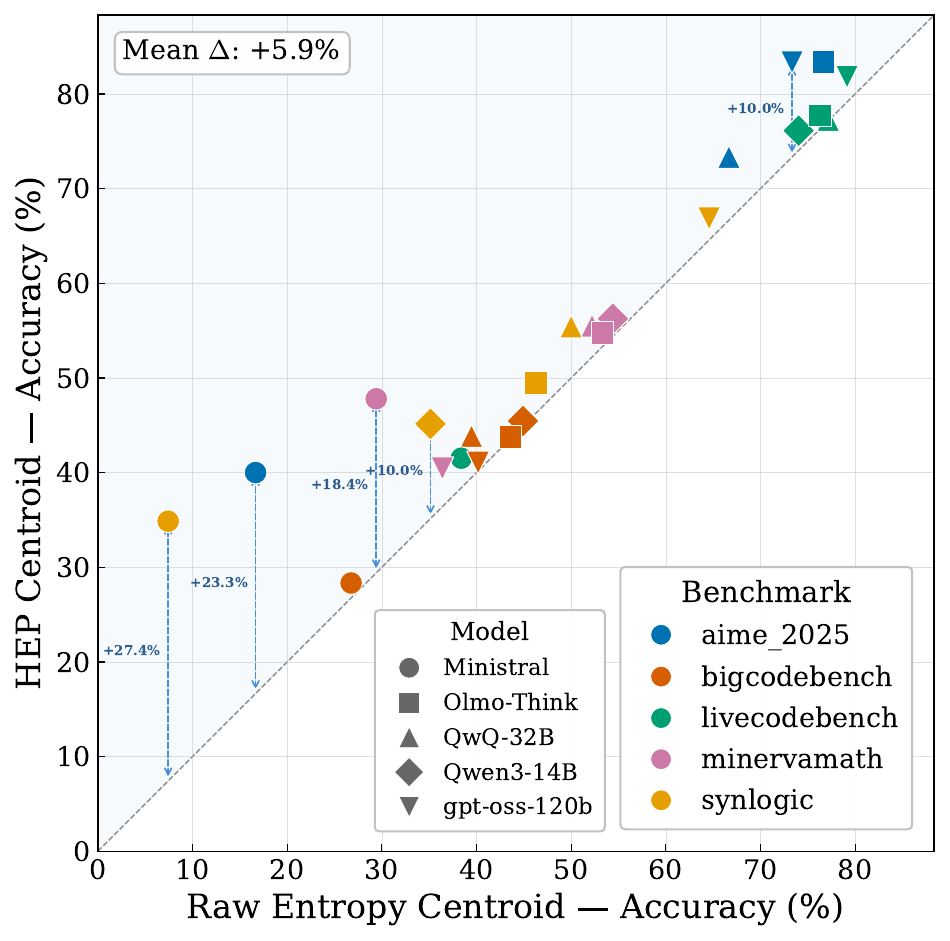}
    \caption{\textbf{HEP Centroid vs. Raw Entropy Centroid.} Each point represents a model-dataset combination. Points above the diagonal indicate that HEP outperforms Raw Entropy. All of the points fall above the diagonal, with a mean improvement of +5.8\%.}
    \label{fig:raw}
\end{figure}

\textbf{Transitioning from token-level raw entropy to a group-level high entropy phase provides an important mechanism to reduce noise during centroid computation.} A core innovation of our work involves proposing the concept of the high entropy phase to represent the model inference state. As discussed in Section \ref{sec:centroid-compute}, we adopt a binary entropy mass approximation to reduce noise. Here, we perform an ablation study on this choice and compare with raw entropy centroid, where we use the raw entropy value of all tokens to compute the centroid. As illustrated in Figure~\ref{fig:raw}, the HEP-based centroid outperforms the raw entropy centroid in all of model-dataset combinations, with a mean accuracy gain of +5.8\% absolutely. The improvement is especially large in lower-accuracy settings, where individual token noise has a stronger influence on centroid computation.

\begin{figure*}[!t] 
  \centering 
  \includegraphics[width=\textwidth]{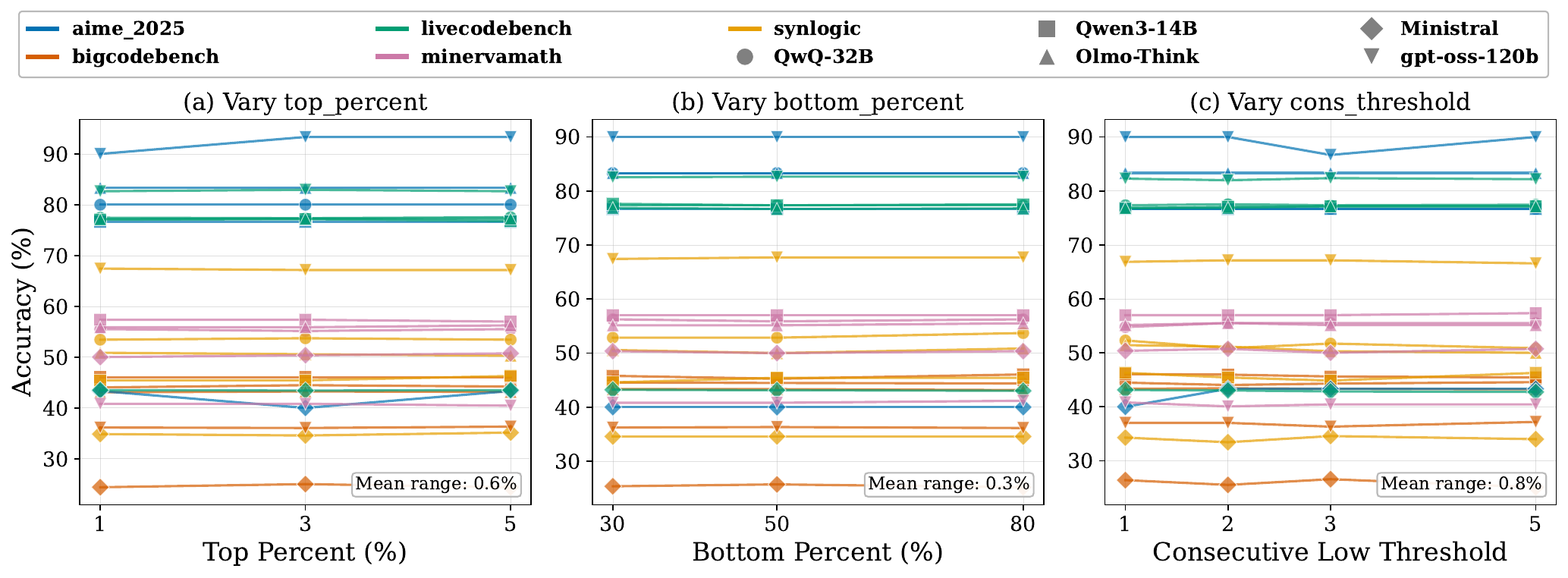} 
  \caption{Hyperparameter robustness of HEP. We vary three parameters independently using a controlled variable approach: (a) the top percent threshold $\theta_{high}$, (b) the bottom percent threshold $\theta_{low}$, and (c) the consecutive low threshold $k$. Each curve represents a model-dataset pair across five models and five benchmarks. All curves remain nearly flat across the tested ranges. The mean performance range is 0.6\%, 0.3\%, and 0.8\% for the three parameters respectively, confirming that HEP is robust to hyperparameter choices.}
  \label{fig:sen}
\end{figure*}

\textbf{The high entropy phase exhibits stable performance and excellent robustness across different hyperparameters.} The definition of the high entropy phase relies on three hyperparameters, specifically the top entropy threshold $\theta_{high}$, the bottom entropy threshold $\theta_{low}$, and the number of continuous bottom entropy tokens $k$ that determine the exit from the high entropy phase. We analyze the hyperparameter robustness of HEP using a controlled variable approach. As shown in Figure~\ref{fig:sen}, model performance remains largely unaffected by variations in parameter values. For all three parameters, the performance difference remains below 1\%, which further confirms HEP is robust to hyperparameter choices. Notably, HEP with $k{=}1$ is not equivalent to raw token-level entropy, due to our binary entropy mass approximation discussed in Section \ref{sec:centroid-compute}. Setting $k > 1$ further prevents isolated low-entropy tokens from prematurely splitting a phase, but the core mechanism remains the same. This explains the comparable performance across different $k$ values. More ablation experiments on filter impact can be found in Appendix~\ref{app:exp}.

\section{\textsc{Scaling Efficiency Analysis}}\label{analysis}

\begin{figure*}[!t]
  \centering 
  \includegraphics[width=\textwidth]{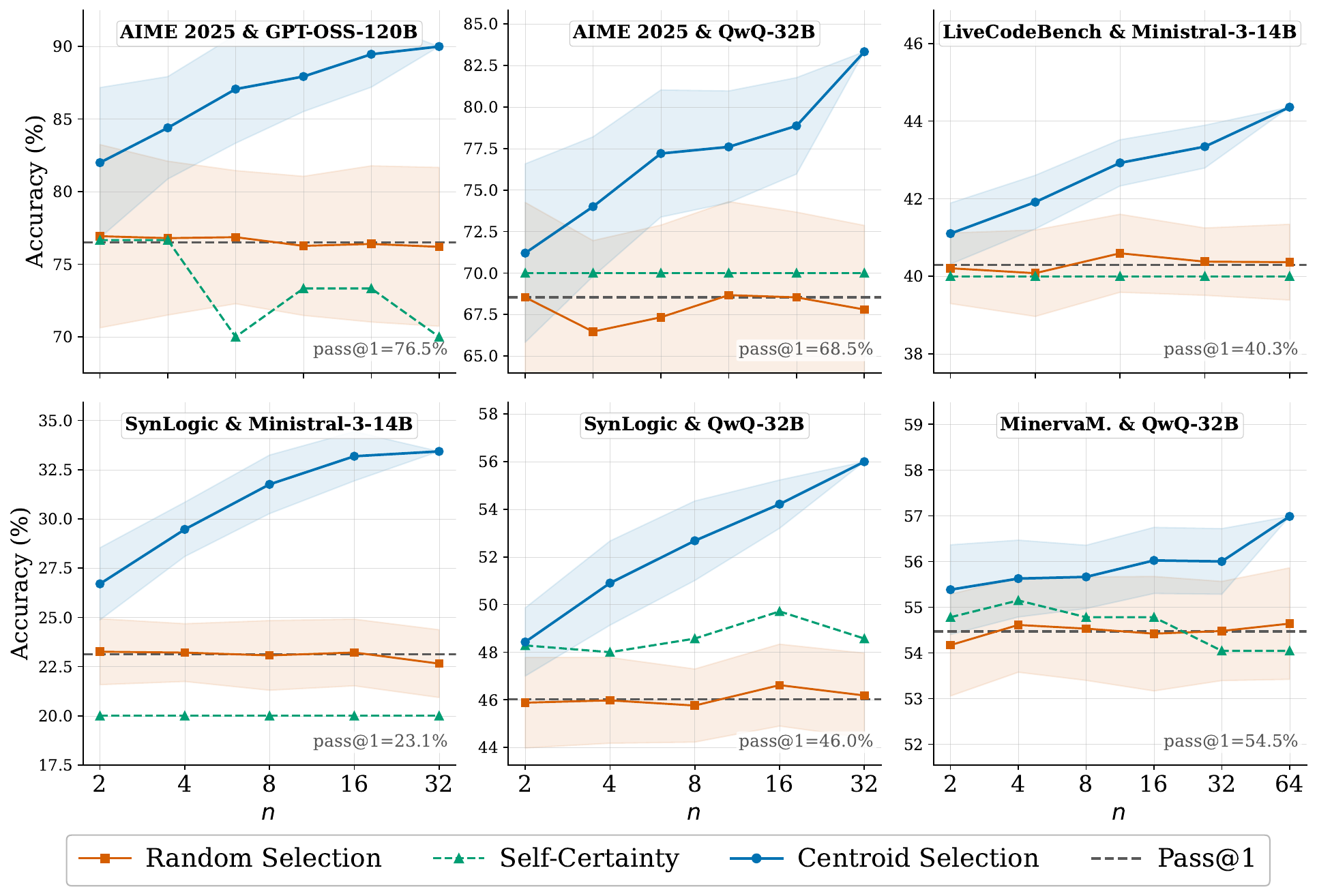} 
  \caption{Test-time scaling performance across models and datasets. Each subplot shows accuracy as a function of the number of sampled trajectories $n$. For each $n$, we repeat the sampling process 50 times. Solid curves denote the mean accuracy and shaded regions denote the standard deviation. Centroid Selection (\textcolor{blue}{blue}) exhibits steady and consistent scaling across all settings without early saturation. Self Certainty (\textcolor[HTML]{2E7D32}{green}) shows unstable scaling with occasional performance drops. Random Selection (\textcolor{orange}{orange}) remains largely flat, close to the Pass@1 baseline (\textcolor{black}{black dashed}).}
  \label{fig:scaling}
\end{figure*}


Test-time scaling cost-performance tradeoff is a key metric for evaluating scaling methods. In our setup, cost and performance manifest as the relationship between the number of sampled responses and accuracy. We compare our method against pass@1, random selection, and the intrinsic reward baseline Self-Certainty.

\textbf{Lowest Centroid exhibits persistent scaling without early saturation.} As shown in Figure~\ref{fig:scaling}, accuracy steadily increases with more sampled responses, without noticeable diminishing returns. In contrast, Self-Certainty shows unstable scaling with occasional performance drops, lacking a consistent upward trend. Comparisons with pass@1 and random baselines further demonstrate the superior scaling behavior of our method. Results span different task domains and models ranging from 14B to 120B, demonstrating the generality of the Lowest Centroid method.

A key advantage of our method is its broad applicability. Lowest Centroid selects a single trajectory based on intrinsic signals without requiring answer labels, making it applicable to any task type, including coding and agentic tasks where methods like majority voting are fundamentally inapplicable due to their reliance on answer tallying. Beyond trajectory selection, the Entropy Centroid can further serve as a reward signal during training, offering wider application potential.

\section{\textsc{Conclusion}}

In this work, we propose to explore the temporal pattern during inference. We define High Entropy Phase (HEP) as the basic unit, a group-level unit that segments consecutive high-entropy tokens into coherent phases. Based on HEP, we define the Entropy Centroid to encode positional information along the trajectory. A lower centroid means the model explores early and converges confidently, indicating higher inference quality. Experiments across different tasks show that the Lowest Centroid outperforms existing baselines, with stable gains from 14B to 480B parameters. We hope the Entropy Centroid inspires further exploration beyond trajectory selection.

\bibliography{colm2026_conference}
\bibliographystyle{colm2026_conference}

\clearpage

\appendix

\section{\textsc{Additional Related Work}}\label{app:related}

\subsection{Intrinsic rewards as metrics for LLM capabilities.}

Exploring intrinsic rewards is essential for achieving self-improvement and continual learning in language models. Current research actively analyzes and models the output distribution. For example, \cite{kang2025scalable,zhao2025learning} calculate the KL divergence between the model distribution and a uniform distribution to serve as an intrinsic reward for self-certainty. Other approaches \citep{fu2025deep,sharma2025think} use output confidence over specific time windows as a signal to enable early stopping for low-quality trajectories and to optimize computational resource allocation during training, making language models more efficient~\citep{zhao-mutis}. Furthermore, \cite{wang2025beyond,hu2026entropy} confirm the significant value of high entropy tokens in model outputs. Additional studies \citep{chuang2023dola,chen2024context,yu2025trajselector,he2025mmboundary,yang2026trust} measure confidence and factuality using hidden states to inform trajectory selectors. However, existing intrinsic reward methods exhibit several limitations. Averaging the complete trajectory of an inference rollout ignores variations in local inference steps. Treating the logits distribution as a fixed signal fails to account for semantic information across different contexts. These methods also struggle to avoid the noise inherent in token-level signals, which ultimately leads to distorted reward signals.

\subsection{Test-time Scaling Methods}

Beyond intrinsic states, researchers explore various unlabeled optimization approaches. Test-time reinforcement learning \citep{zuo2025ttrl,zhou2025evolving} utilizes model voting results to achieve self-improvement. Incorporating large language models for evaluation during test time is another widely adopted practice \citep{zheng2023judging,chan2023chateval,wang2025mcts}. Researchers often employ multi-agent debate \citep{yang2025mars, he2025medtutor} to improve verification accuracy within this context. These evaluation methods require external models, which incur additional overhead and introduce model-based noise. Monte Carlo tree search is frequently applied to expand the solution space of model outputs during test time \citep{misaki2025wider,li2025solverllm}. This search enhances model performance through systematic exploration and backtracking mechanisms. Nevertheless, traversing the solution space demands substantial computational resources and exhibits diminishing returns \citep{ghosal2025does}, making these search methods unsuitable for practical production environments.


\section{\textsc{Additional Experiment Details}}\label{app:exp}

\subsection{Additional Experiment Results}

We present more majority voting results here for readers' reference.

\begin{table}[htbp]
      \centering
      \renewcommand{\arraystretch}{1.2}
      \setlength{\tabcolsep}{3.5pt}
      \resizebox{\textwidth}{!}{
      \begin{tabular}{l *{9}{l}}
          \toprule[1.2pt]

          \textbf{Model}
          & \multicolumn{3}{c}{\textbf{AIME25}} &
  \multicolumn{3}{c}{\textbf{Minerva}}
          & \multicolumn{3}{c}{\textbf{Synlogic}} \\

          \cmidrule(lr){2-4} \cmidrule(lr){5-7}
  \cmidrule(lr){8-10}

       Acc $(\%,\uparrow)$   & \scriptsize Pass@1 & \scriptsize Major. Voting &
  \tiny Lowest Cent.
          & \scriptsize Pass@1 & \scriptsize Major. Voting &
  \tiny Lowest Cent.
          & \scriptsize Pass@1 & \scriptsize Major. Voting &
  \tiny Lowest Cent. \\

          \midrule[0.8pt]

          Olmo-3.1 & 75.7 & 80.0\gain{4.3} & \superbold{86.7}\gain{11.0} & 55.0 & \superbold{56.3}\gain{1.3} & 55.9\gain{0.9} & 42.1 & 46.9\gain{4.8} & \superbold{50.0}\gain{7.9} \\
          
          Ministral-3 & 27.5 & \superbold{40.0}\gain{12.5} & \superbold{40.0}\gain{12.5} & 46.7 & \superbold{55.2}\gain{8.5} & 50.4\gain{3.7} & 23.1 & \superbold{41.1}\gain{18.0} & 33.4\gain{10.3} \\
          
          Qwen3-14B & 71.4 & \superbold{80.0}\gain{8.6} & 76.7\gain{5.3} & 56.3 & 56.6\gain{0.3} & \superbold{58.1}\gain{1.8} & 44.4 & \superbold{49.1}\gain{4.7} & 45.4\gain{1.0} \\
          
          GPT-OSS & 76.5 & \superbold{90.0}\gain{13.5} & \superbold{90.0}\gain{13.5} & 38.6 & 39.7\gain{1.1} & \superbold{40.1}\gain{1.5} & 63.1 & \superbold{68.9}\gain{5.8} & 67.4\gain{4.3} \\
          
          QWQ-32B & 68.5 & 80.0\gain{11.5} & \superbold{83.3}\gain{14.8} & 54.5 & 56.3\gain{1.8} & \superbold{57.0}\gain{2.5} & 46.0 & 48.0\gain{2.0} & \superbold{56.0}\gain{10.0} \\

          \bottomrule[1.2pt]
          
      \end{tabular}}
      \caption{For short-answer tasks where majority voting is applicable, we compare Lowest Centroid against majority voting. The results show that the two methods achieve comparable performance.}
      \label{tab:majority_voting}
  \end{table}

\subsection{Additional Ablation Results}

\textbf{Outlier filtering primarily addresses extreme cases and exerts a minimal impact on overall performance.} As previously mentioned, certain inference trajectories suffer from early stopping and repetition issues.

To verify the impact of outlier filtering on experimental results, we compare the accuracy before and after filtering. We conduct this comparison across different model–dataset combinations to ensure rigor. For most models, the effect of outlier filtering on performance is less than 1\%. We validate this on every model–dataset pair, and the results are presented in Table~\ref{tab:filter_impact}. As shown, outlier filtering has a negligible impact on model performance, further confirming that our performance gains primarily stem from the design of the entropy centroid. Outlier filtering serves solely to remove trajectories caused by repetition or early stopping. The filtering effect in AIME 2025 appears to have a significant impact on the data, but this is because AIME only has 30 questions. While affecting just one question would result in a 3.3\% increase in accuracy, in reality, only about 1.5 questions are affected on average in AIME.

\begin{table}[!t]
      \centering
      \renewcommand{\arraystretch}{1.2}
      \setlength{\tabcolsep}{3.5pt}
      \begin{tabular}{cccccc}
          \toprule[1.2pt]
          \textbf{Model  $\Delta (\% ,\downarrow)$  } & \scriptsize \textbf{Bigcode} & \scriptsize \textbf{Livecode} & \scriptsize \textbf{Minerva} & \scriptsize \textbf{Synlogic} & \scriptsize \textbf{AIME} \\

          \midrule[0.8pt]

          Olmo-3.1-32B-Think & 0.54 & 0.09 & 1.32 & 2.29 & 3.33 \\
          Qwen3-14B & 0.21 & 2.26 & 1.10 & 3.37 & 6.67 \\
          QWQ-32B & 0.98 & 0.47 & 0.44 & 1.09 & 6.67 \\
          Ministral-3-14B-Instruct & 0.00 & 3.15 & 2.28 & 1.66 & 4.67 \\
          GPT-OSS-120B & 0.09 & 1.02 & 0.81 & 0.91 & 4.67 \\

          \bottomrule[1.2pt]
      \end{tabular}
      \caption{ Accuracy changes before and after outlier filtering across different datasets and models. \textbf{For each dataset–model combination, we evaluate 5 distinct high-entropy phase configurations and report the average accuracy difference as the filtering impact.}}
      \label{tab:filter_impact}

  \end{table}

\subsection{Evaluation Details}

Our inference framework is built on vLLM \citep{kwon2023efficient}. The evaluation for each domain task follows well-established references. For mathematical datasets, we adopt the evaluation logic from LIMO \citep{ye2025limoreasoning} and Qwen \citep{yang2024qwen25mathtechnicalreportmathematical}, relying exclusively on rule-based rewards without any model-based judgment. All other datasets are evaluated using the original evaluation code provided by their respective authors.

To ensure evaluation stability and experimental rigor, we first run a unified evaluation pass during the inference stage to generate an \textbf{evaluation cache}. All subsequent evaluation methods read directly from this cache, eliminating potential inconsistencies introduced by non-deterministic evaluation procedures.

Table~\ref{parameter1} and~\ref{parameter2} summarize the inference hyperparameters for each dataset. We follow the officially recommended configurations and make dataset-specific adjustments according to available computational resources. Since $\tau^2$-Bench is a multi-turn agent benchmark, its parameter configuration differs from those of the other single-turn datasets.

\begin{table}[!t]
\centering
\resizebox{\textwidth}{!}{%
\begin{tabular}{llllllll}
\toprule
\multicolumn{2}{c}{\textbf{BIGCODE}} & \multicolumn{2}{c}{\textbf{LIVECODE}} & \multicolumn{2}{c}{\textbf{MATH}} & \multicolumn{2}{c}{\textbf{LOGIC}} \\
\midrule
max\_token & 16384  & max\_tokens &  32768   & max\_tokens & 32768 & max\_tokens & 32768      \\
batch\_size   & 16  & batch\_size   & 256   & batch\_size    & 128  & batch\_size         & 64      \\
temperature & 0.7 & temperature & 0.7 & temperature & 0.7& temperature      & 1.0      \\
n(\#trajectories) & 32 & n(\#trajectories) & 32 & n(\#trajectories) & 64 & n(\#trajectories) & 64 \\
\bottomrule
\end{tabular}
}%
\caption{Inference Hyperparameter Settings}
\label{parameter1}
\end{table}

\begin{table}[!t]
\centering

\begin{tabular}{ll}
\toprule
\multicolumn{2}{c}{\textbf{$\tau^2$-Bench}}\\
\midrule
AGENT\_MAX\_TOKENS & 8192  \\
USER\_MAX\_TOKENS & 2048 \\
MAX\_STEPS   & 200    \\
AGENT\_TEMP & 0.7    \\
USER\_TEMP & 0.7    \\
SEED & 300 \\
n(\#trajectories) & 64  \\
\bottomrule
\end{tabular}

\caption{$\tau^2$-Bench Inference Hyperparameter Settings}
\label{parameter2}
\end{table}

\subsection{Dataset Details}

\textbf{AIME 2025}~\citep{aime25} is the 2025 edition of the American Invitational Mathematics Examination, a prestigious invitational competition targeting high-performing high school students. It comprises 30 problems across two sessions (AIME I and AIME II, each with 15 problems), covering algebra, geometry, number theory, combinatorics, and probability. Each problem requires an integer answer between 000 and 999 and demands creative, multi-step mathematical reasoning well beyond standard curricula. We use all 30 problems as our test set.

\textbf{Minerva Math}~\citep{lewkowycz2022solving} is a mathematical reasoning benchmark derived from the Minerva evaluation suite, consisting of undergraduate-level STEM problems that require multi-step quantitative reasoning. Problems span topics such as calculus, linear algebra, probability, and differential equations, and models must produce detailed solutions with precise numerical or symbolic answers. Its test set contains 272 unique problems.

\textbf{Synlogic}~\citep{liu2025synlogic} is a synthetic logical reasoning benchmark in which problems are algorithmically generated to ensure verifiable ground-truth answers. It evaluates language models' capacity for formal deductive reasoning across multiple categories of logical inference. We use the Val-hard subset, which contains the most challenging problems requiring complex, multi-step logical deductions.

\textbf{LiveCodeBench}~\citep{jain2024livecodebench} is a contamination-free code generation benchmark that continuously curates new problems from competitive programming platforms such as LeetCode, AtCoder, and Codeforces. Unlike static benchmarks, its rolling release strategy mitigates data leakage. We use release v6, which evaluates code generation capabilities where models must produce correct, executable solutions given natural language problem descriptions and test cases.

\textbf{BigCodeBench}~\citep{zhuo2024bigcodebench} is a practical programming benchmark comprising 1,140 function-level tasks that span 7 domains and require the use of 139 diverse libraries. Unlike competition-style benchmarks, it emphasizes real-world programming scenarios demanding practical API knowledge and compositional reasoning. We evaluate using the instruct mode (v0.1.4), where models receive natural language task descriptions and must generate standalone, executable functions.

\textbf{$\tau$2-bench}~\citep{barres2025tau} is an agentic benchmark designed to evaluate language model agents' ability to handle realistic, multi-turn customer service interactions, where agents must invoke tools and adhere to domain-specific policies to resolve user requests. We use the Test subset containing 100 items spanning three domains: airline (20 items), retail (40 items), and telecom (40 items), collectively covering diverse tool-use and policy-following scenarios.

\section{\textsc{Implementation Details}}\label{app:method}

\definecolor{PromptFrameBlue}{HTML}{4A6984} 
\definecolor{PromptBackBlue}{HTML}{F8F9FB}

\definecolor{PromptFrameWarm}{HTML}{9E8576} 
\definecolor{PromptBackWarm}{HTML}{FCFBF9}

\newtcolorbox{systemprompt}[2][]{
    enhanced,
    breakable,
    pad at break=2mm,
    bottomrule at break=1pt,
    toprule at break=1pt,
    colframe=PromptFrameBlue, 
    colback=PromptBackBlue,   
    coltitle=white,           
    fonttitle=\sffamily\fontseries{bx}\selectfont\small, 
    fontupper=\small,          
    arc=3pt,                  
    boxrule=0.8pt,            
    left=8pt, right=8pt,     
    top=6pt, bottom=6pt,      
    boxsep=2pt,
    titlerule=0mm,           
    toptitle=3pt, bottomtitle=3pt,
    title={#2},               
    #1                        
}

\subsection{Additional Method Details}

\begin{figure}[htbp] 
  \centering 
  \includegraphics[width=\textwidth]{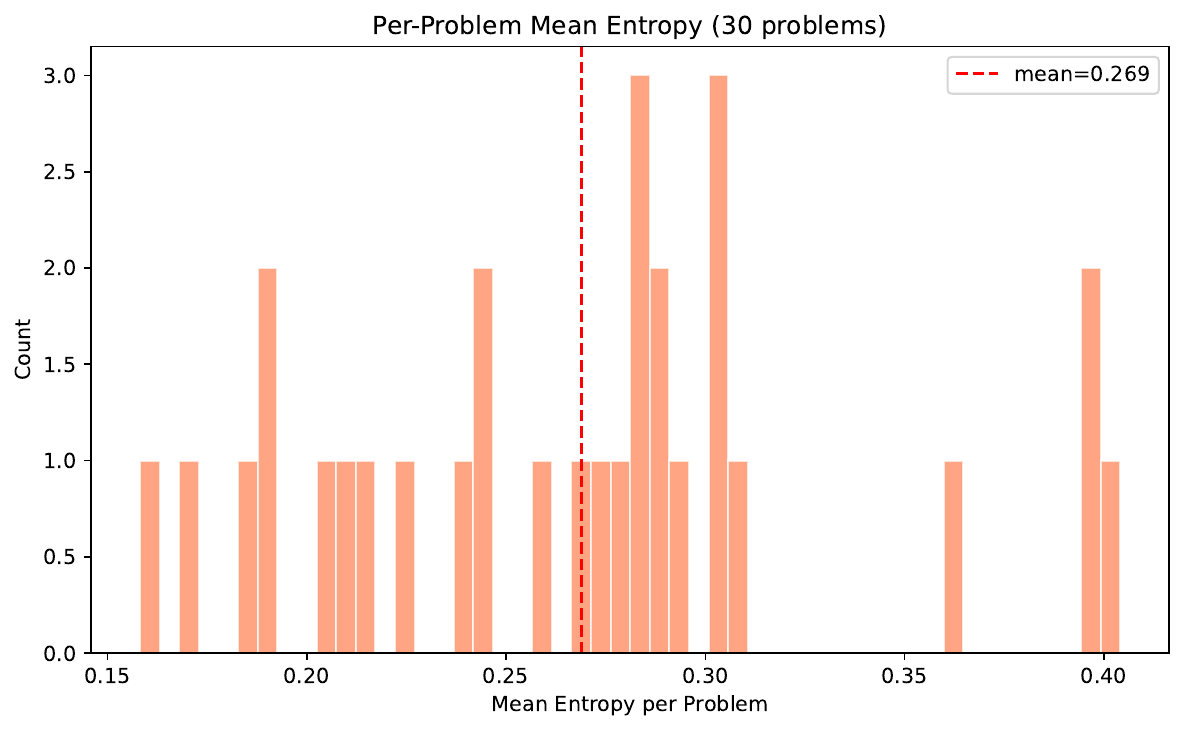} 
  \caption{The mean Entropy values differ across different problems, which verifies that using absolute value as a threshold is not feasible.}
  \label{fig:ent}
\end{figure}

\subsubsection{Percentile-based Threshold for HEP Detection}

We use a percentile-based threshold rather than an absolute entropy value to define HEP. This design choice is motivated by both theoretical and practical considerations.

First, entropy distributions vary significantly across models and datasets, as shown in Figure~\ref{fig:ent}. A fixed absolute threshold cannot generalize across these different distributions. For example, if the threshold is set to 3 but no token in a trajectory exceeds an entropy of 3, the resulting centroid defaults to zero, producing a meaningless signal.

Second, the core of our method is to characterize entropy variation across the trajectory, i.e., the transitions between inference states. An absolute threshold cannot guarantee a sufficient number of high-entropy tokens for stable state characterization. In contrast, a percentile-based threshold adapts to the entropy distribution of each trajectory, ensuring that the relative fluctuation patterns are consistently captured regardless of the model or dataset.

\subsubsection{Discover LLM Emotion Pattern for Self Improvement}

The perspective of emotion originates from~\cite{llyaemo2025}. Self-improvement has emerged as a central topic in current research, with state-of-the-art models increasingly emphasizing iterative self-refinement as a key component of their development. The question of how to achieve model self-improvement has evolved from an open research problem into a widely adopted training paradigm. In this context, the entropy centroid we propose offers a promising intrinsic reward signal and value function for self-improvement. Beyond serving as an objective score that reflects inference quality, the entropy centroid provides a deeper characterization of the model's internal reasoning dynamics. We move beyond token-level entropy, which suffers from substantial noise and limited interpretability, and introduce a new paradigm for characterizing inference states — one that yields dense, high-quality signals for model training.

\subsection{Future Work}

In this work, we focus on applying the Entropy Centroid as an intrinsic reward for test-time scaling. However, the underlying modeling framework, characterizing inference states through HEP and encoding their temporal structure via centroids, is not restricted to this setting. The only requirement is access to the model's token-level entropy during generation, which is readily available in any autoregressive decoding process. This makes the approach naturally extensible to other scenarios. For example, the Entropy Centroid could serve as an auxiliary reward signal in on-policy training, providing a lightweight and annotation-free complement to external rewards. More broadly, the temporal patterns captured by HEP may offer useful diagnostics for analyzing model behavior during long-form generation, multi-turn dialogue, or agentic workflows. We leave the exploration of these directions to future work.

\subsection{Prompt Design}

When setting the prompt, we generally align it with the default prompt template of the dataset to ensure the rigor and fairness of the evaluation. The specific prompts for different datasets are as follows:

\begin{systemprompt}{MATH DATASET}

\texttt{SYSTEM:} You are a math expert. Please solve problems step by step and put your final answer within \texttt{boxed\{\}}.

\vspace{1em}
\texttt{USER:} Problem: \texttt{\{problem\}}.

\end{systemprompt}

\vspace{1em}

\begin{systemprompt}{SYNLOGIC DATASET}

\texttt{SYSTEM:} You are a logic puzzle solving expert. Analyze the problem carefully, think step by step, and provide your final answer within \texttt{<answer>} and \texttt{</answer>} tags.

\vspace{1em}
\texttt{USER:} \texttt{\{prompt\}} (Extracting the prompt field from the HuggingFace dataset).

\end{systemprompt}

\vspace{1em}

\begin{systemprompt}{BIGCODEBENCH DATASET}

\texttt{SYSTEM:} You are an expert Python programmer. Write clean, efficient, and correct Python code to solve the given task. Include all necessary imports and ensure the code is complete and executable.
\vspace{1em}

\texttt{USER:} 
\begin{verbatim}
    Task:
    {instruct_prompt}
\end{verbatim}

\end{systemprompt}

\vspace{1em}

\begin{systemprompt}{LIVECODEBENCH DATASET}

\texttt{SYSTEM:} You are an expert Python programmer. You will be given a question (problem specification) and will generate a correct Python program that matches the specification and passes all tests. Wrap your final solution in \verb|```| \texttt{python} \verb|```| code blocks.

\vspace{1em}
\texttt{USER:} (with starter\_code, LeetCode Style)

\begin{verbatim}
  ### Question:
  {question_content}
  
  ### Format: 
  You will use the following starter code to write the solution...
  ```python
  {starter_code}
  ```
  
  ### Answer: 
  (use the provided format with backticks)
\end{verbatim}

\vspace{1em}
\texttt{USER:} (without starter\_code, Codeforces/AtCoder Style)

\begin{verbatim}
  ### Question:
  {question_content}
  
  ### Format:
  Read the inputs from stdin, solve the problem and write the answer to stdout...
  ```python
  # YOUR CODE HERE
  ```
  
  ### Answer:
  (use the provided format with backticks)

\end{verbatim}
 
\end{systemprompt}

\vspace{1em}

\section{\textsc{Case Study}}\label{app:case}

\definecolor{PromptFrameWarm}{HTML}{9E8576} 
\definecolor{PromptBackWarm}{HTML}{FCFBF9}

\newtcolorbox{case}[2][]{
    enhanced,
    breakable,
    pad at break=2mm,
    bottomrule at break=1pt,
    toprule at break=1pt,
    colframe=PromptFrameWarm,
    colback=PromptBackWarm,   
    coltitle=white,           
    fonttitle=\sffamily\bfseries\small, 
    fontupper=\small,                   
    arc=3pt,                  
    boxrule=0.8pt,            
    left=8pt, right=8pt,      
    top=6pt, bottom=6pt,      
    boxsep=2pt,
    titlerule=0mm,            
    toptitle=3pt, bottomtitle=3pt,
    title={#2},               
    #1                        
}

We provide detailed inference trajectories for reference. For each problem, we select a correct trajectory with a low centroid and an incorrect trajectory with a high centroid. These concrete examples illustrate how the model's inference state evolves in each case, offering a more intuitive understanding of how the Entropy Centroid reflects the reasoning process.

\subsection{Code Generation Trajectory}

We select a task in Livecodebench and visualize the entire inference process for correct and incorrect trajectories. Figure~\ref{fig:code_wrong} shows the wrong trajectory, while Figure~\ref{fig:code_correct} shows the correct trajectory.

\begin{figure}[p]
    \centering
    \includegraphics[width=\textwidth, page=1]{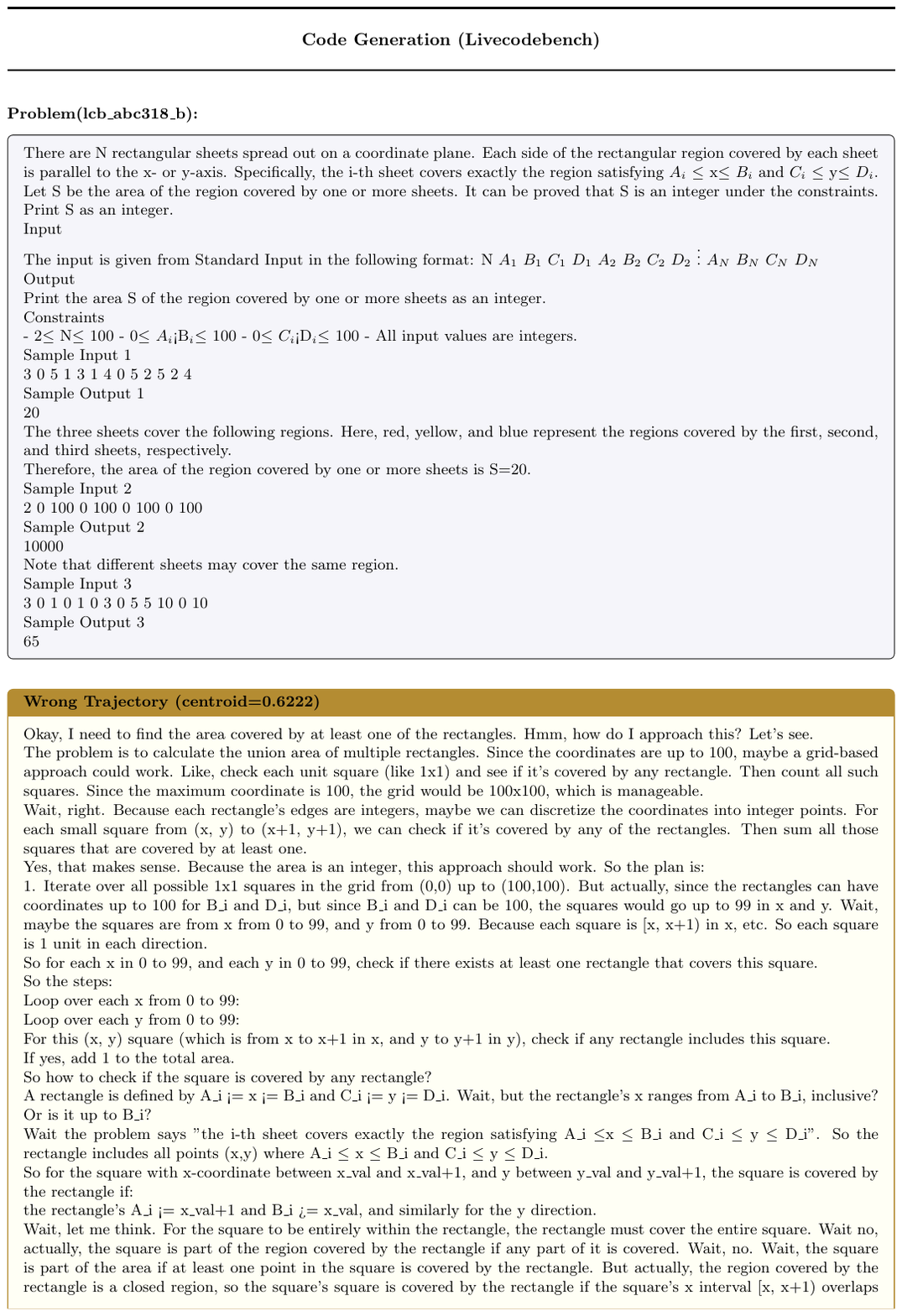}
    \caption{Code Generation Task Livecodebench Example. Wrong Trajectory with High Centroid}
    \label{fig:code_wrong}
  \end{figure}

\begin{figure}[p]
    \ContinuedFloat
    \centering
    \includegraphics[width=\textwidth, page=2]{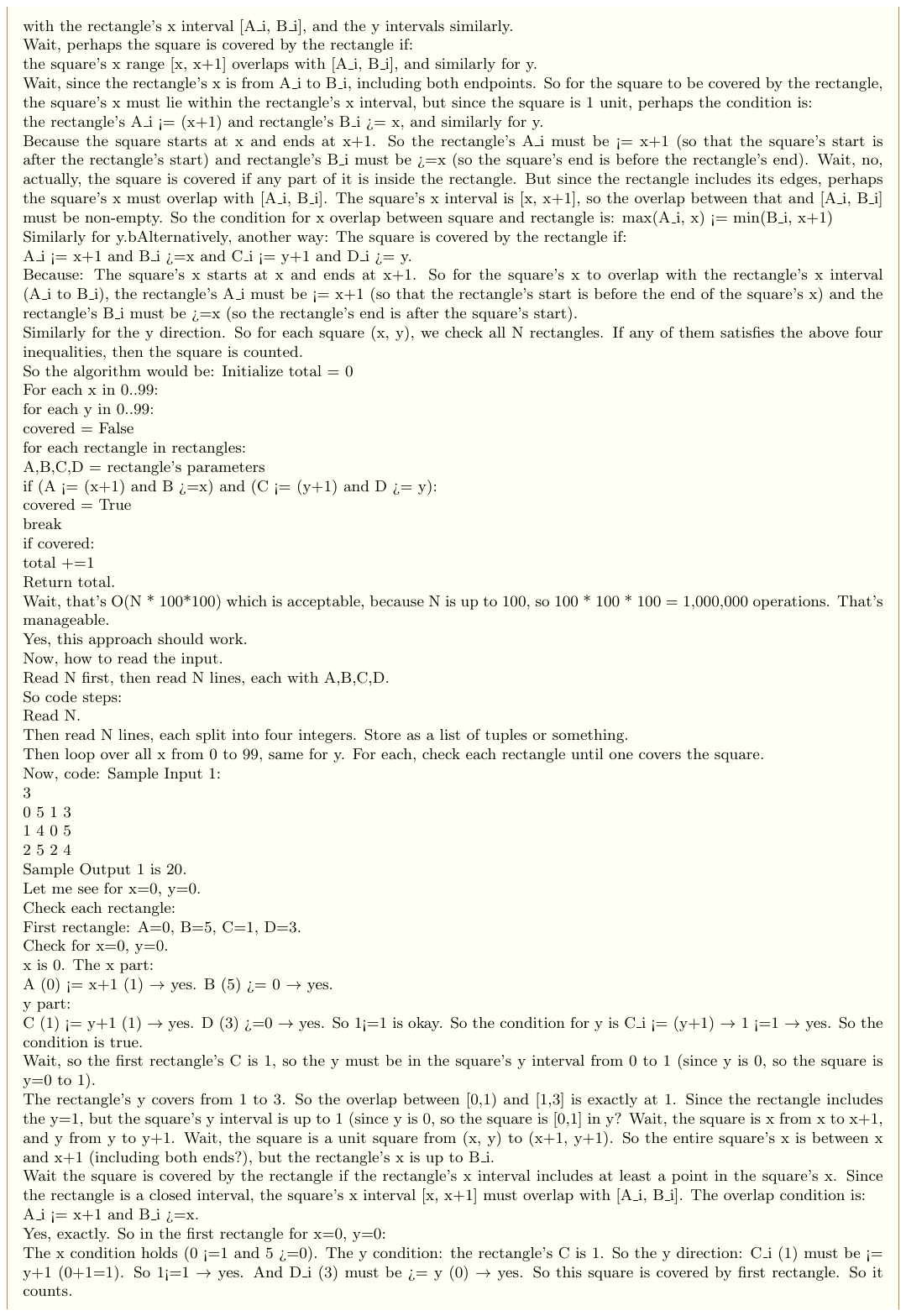}
    \caption{Code Generation Task Livecodebench Example. Wrong Trajectory with High Centroid}
  \end{figure}

\begin{figure}[p]
    \ContinuedFloat
    \centering
    \includegraphics[width=\textwidth, page=3]{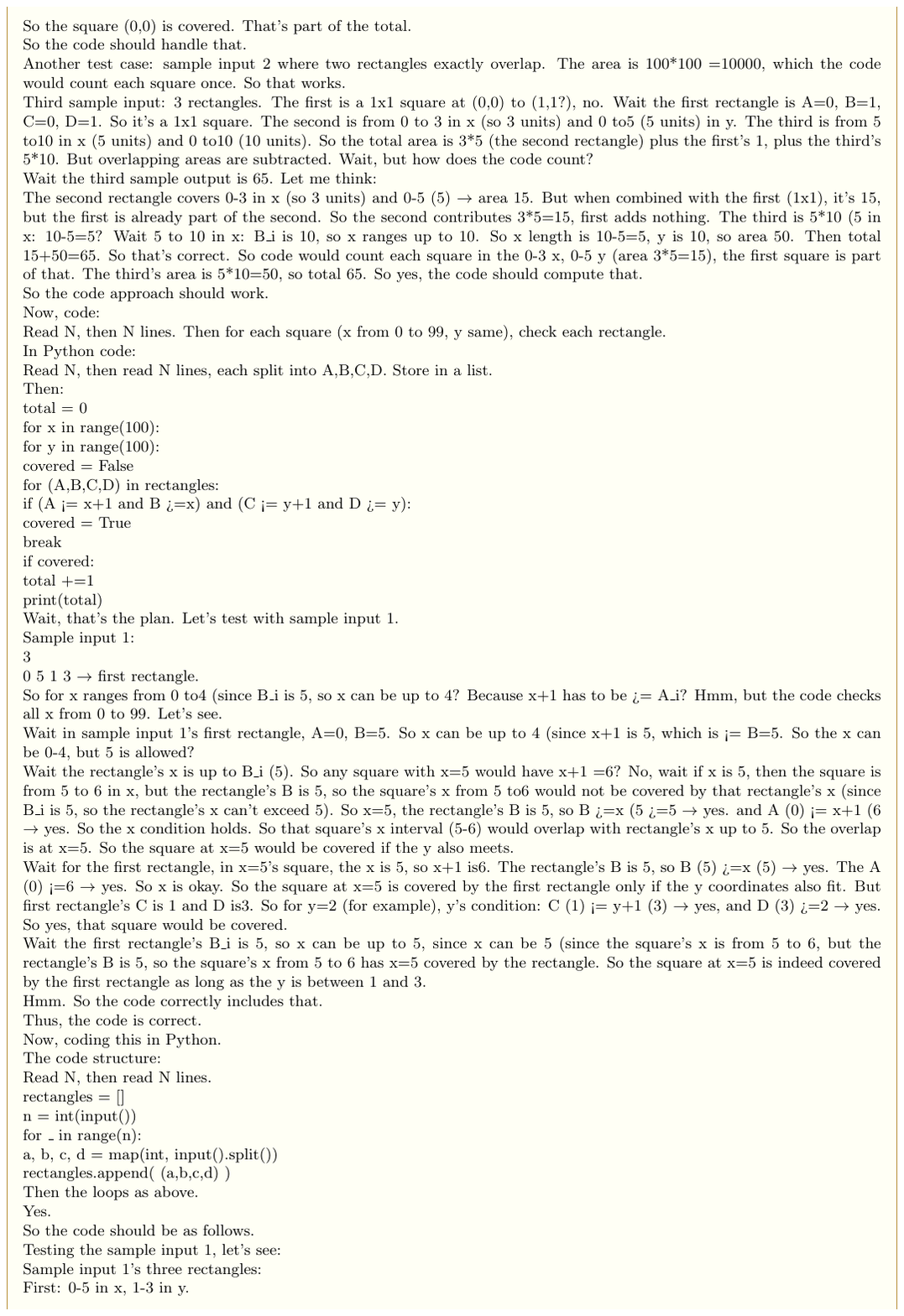}
    \caption{Code Generation Task Livecodebench Example. Wrong Trajectory with High Centroid}
  \end{figure}

\begin{figure}[p]
    \centering
    \includegraphics[width=\textwidth, page=4]{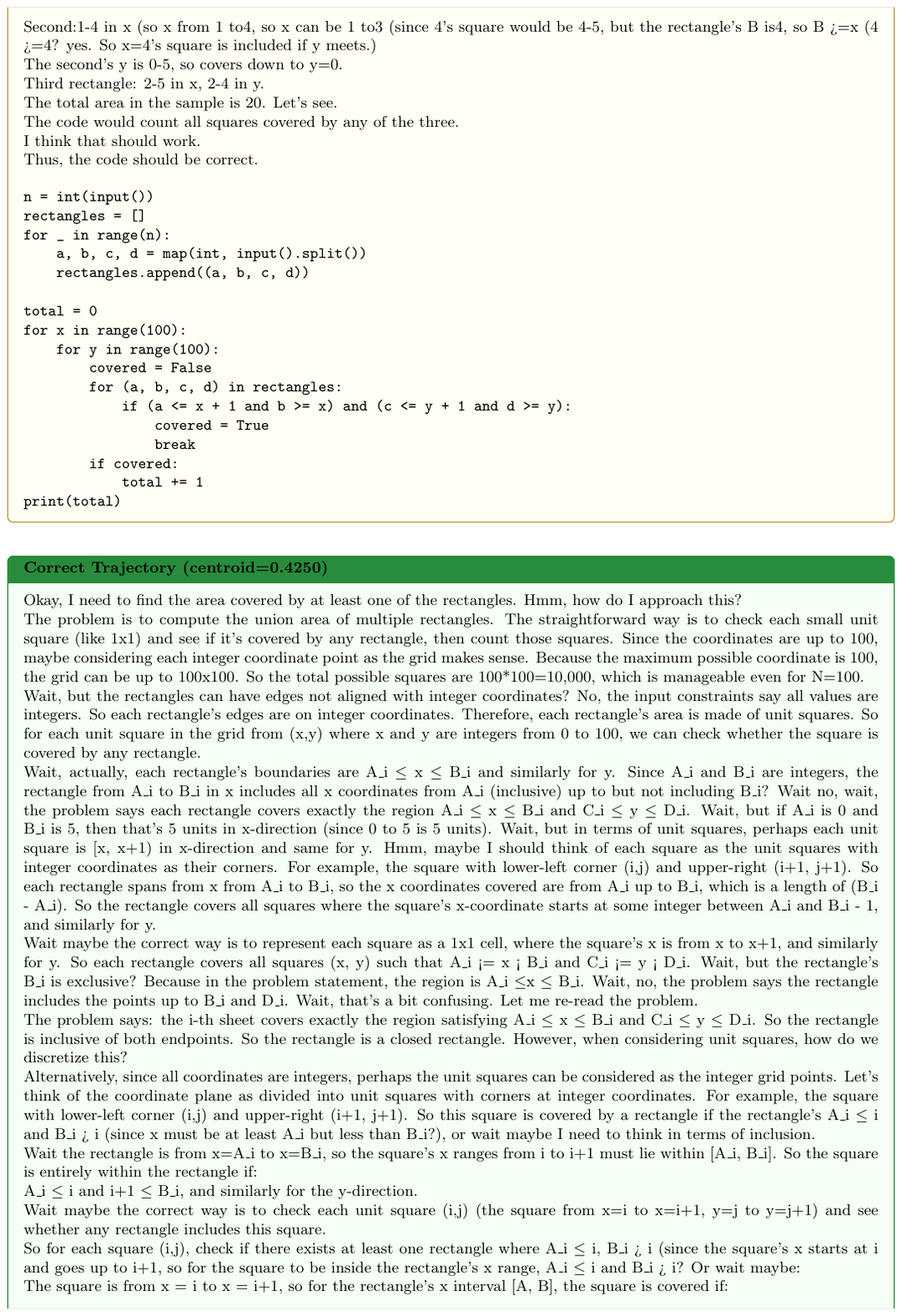}
    \caption{Code Generation Task Livecodebench Example. Correct Trajectory with Low Centroid}
    \label{fig:code_correct}
  \end{figure}

\begin{figure}[p]
    \ContinuedFloat
    \centering
    \includegraphics[width=\textwidth, page=5]{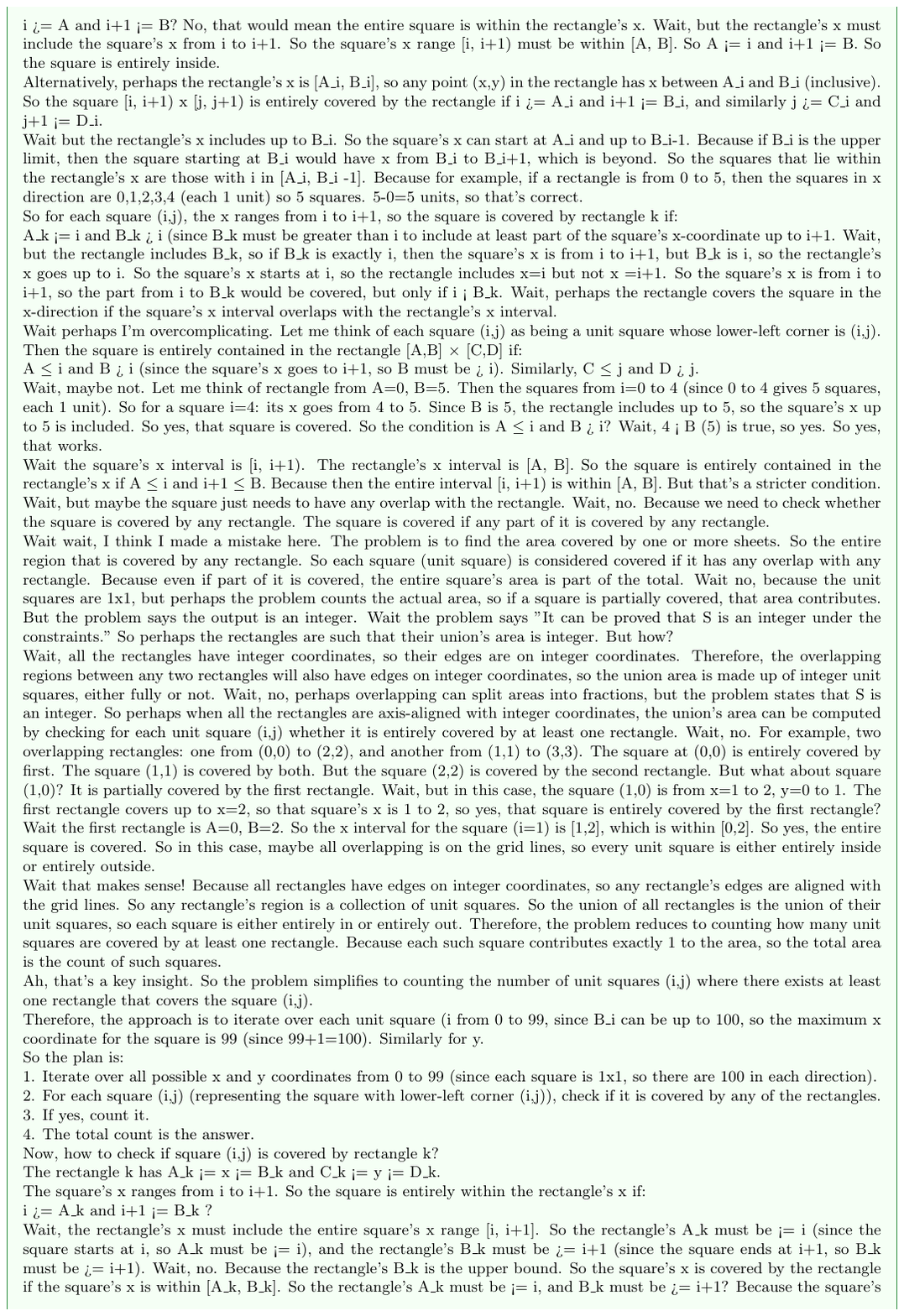}
    \caption{Code Generation Task Livecodebench Example. Correct Trajectory with Low Centroid}
  \end{figure}

\begin{figure}[p]
    \ContinuedFloat
    \centering
    \includegraphics[width=\textwidth, page=6]{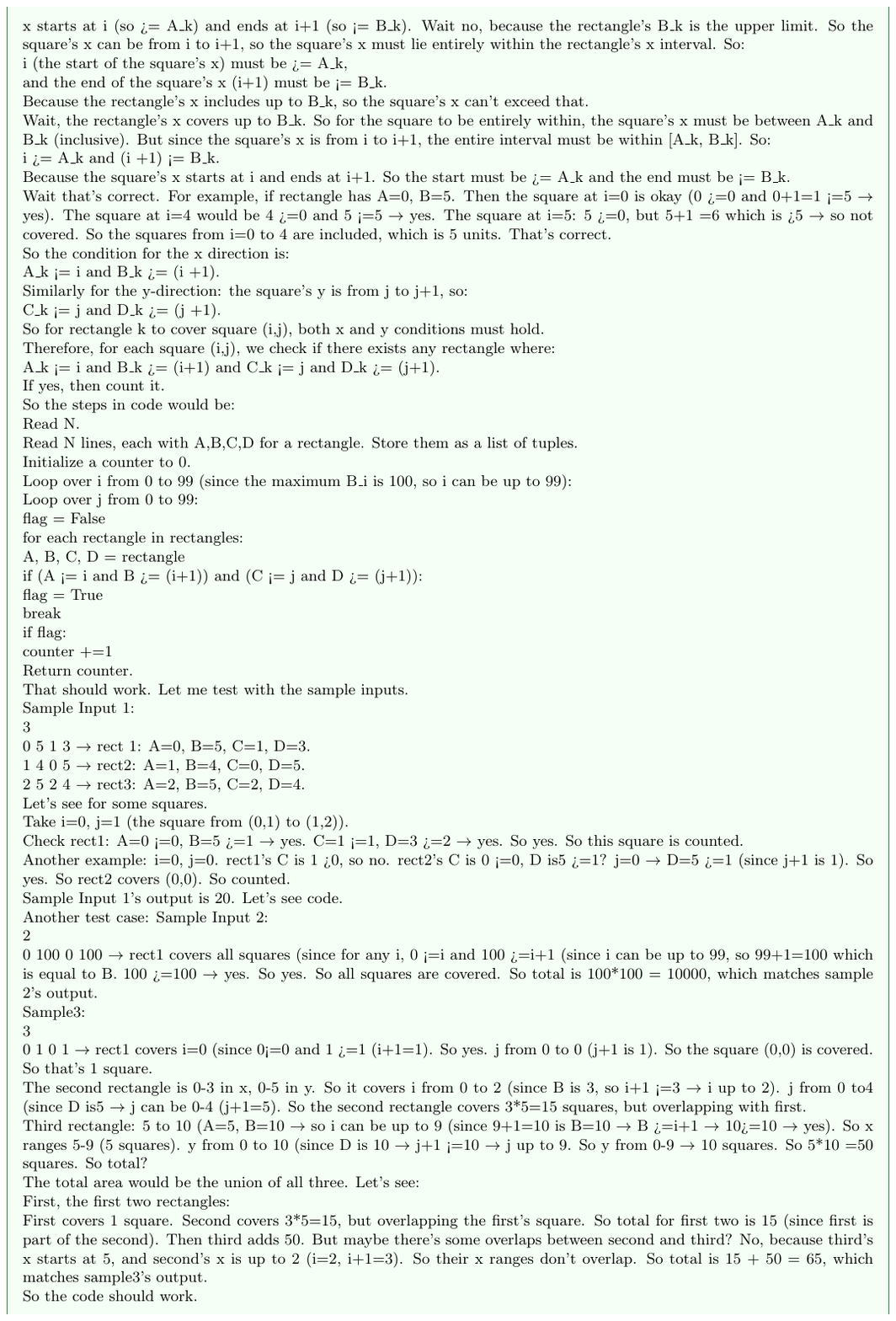}
    \caption{Code Generation Task Livecodebench Example. Correct Trajectory with Low Centroid}
  \end{figure}

\begin{figure}[p]
    \ContinuedFloat
    \centering
    \includegraphics[width=\textwidth, page=7]{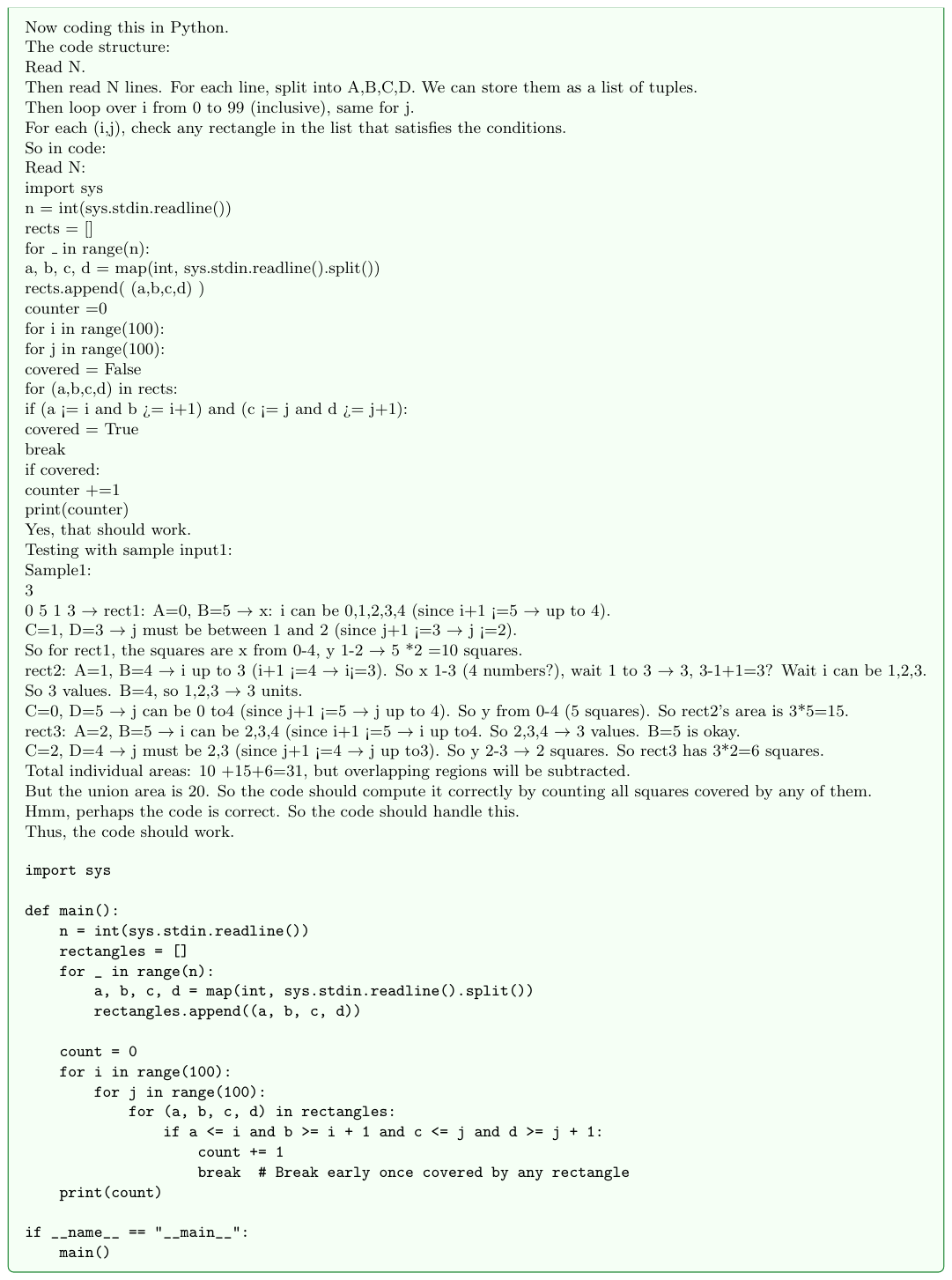}
    \caption{Code Generation Task Livecodebench Example. Correct Trajectory with Low Centroid}
  \end{figure}

\subsection{Logical Reasoning Trajectory}

\begin{figure}[p]
    \centering
    \includegraphics[width=\textwidth, page=1]{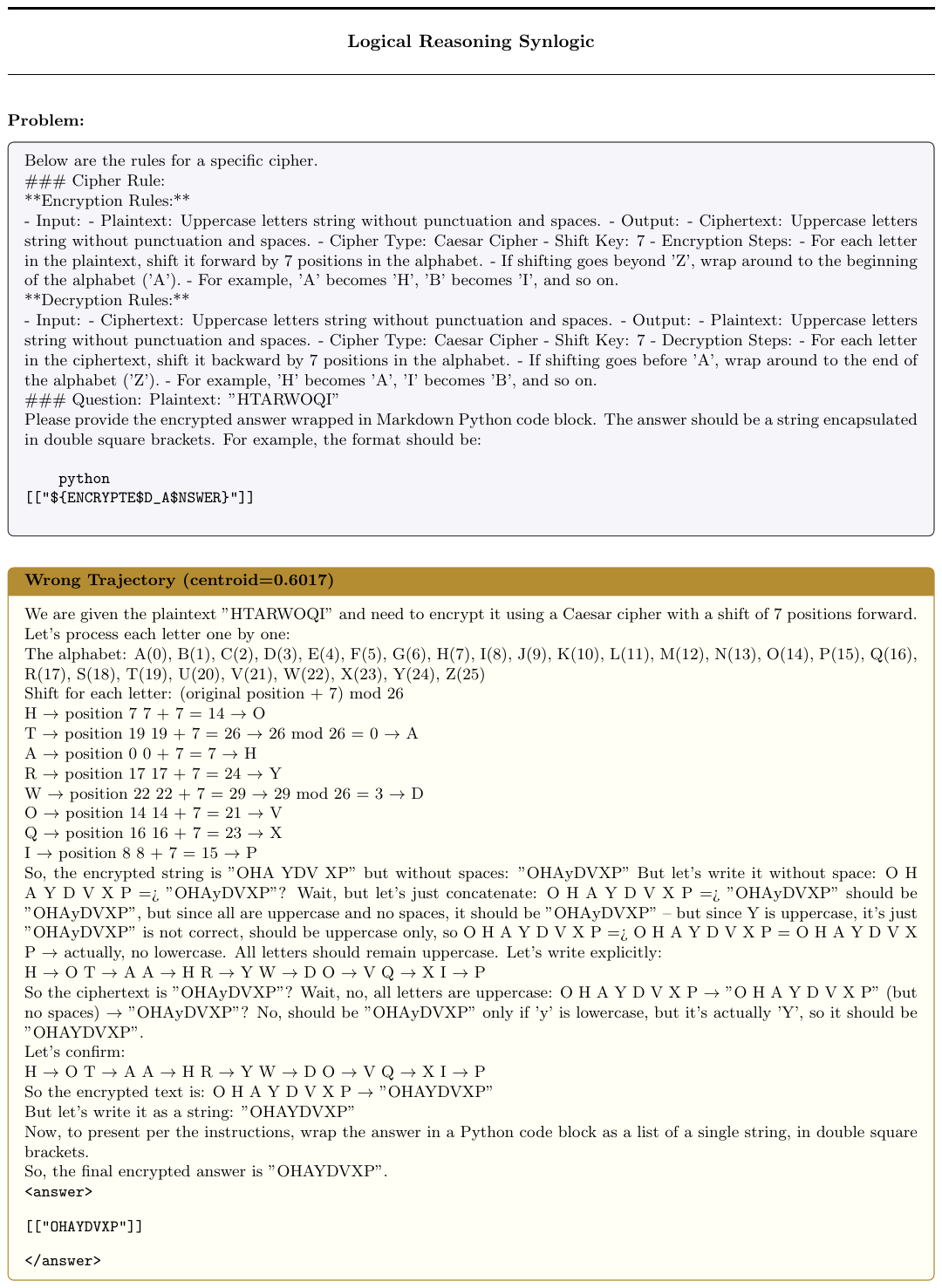}
    \caption{Logic Reasoning Task Synlogic Example. Wrong Trajectory with High Centroid}
    \label{fig:logic_wrong}
  \end{figure}

\begin{figure}[p]
    \centering
    \includegraphics[width=\textwidth, page=2]{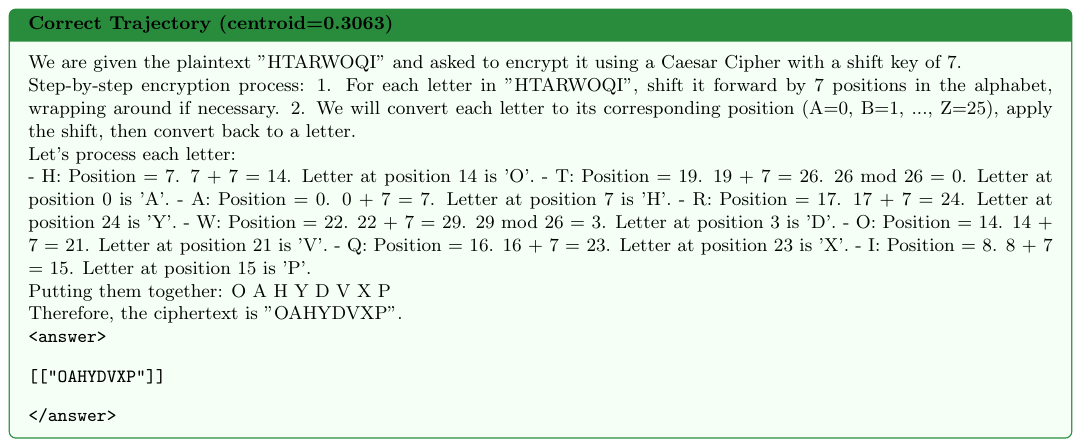}
    \caption{Logic Reasoning Task Synlogic Example. Correct Trajectory with Low Centroid}
    \label{fig:logic_correct}
  \end{figure}

We select a task in Synlogic and visualize the entire inference process for correct and incorrect trajectories. Figure~\ref{fig:logic_wrong} shows the wrong trajectory, while Figure~\ref{fig:logic_correct} shows the correct trajectory.

\subsection{Mathematics Inference Trajectory}

We select a task in AIME 2025 and visualize the entire inference process for correct and incorrect trajectories. Figure~\ref{fig:math_wrong} shows the wrong trajectory, while Figure~\ref{fig:math_correct} shows the correct trajectory.

 \begin{figure}[p]
    \centering
    \includegraphics[width=\textwidth, page=1]{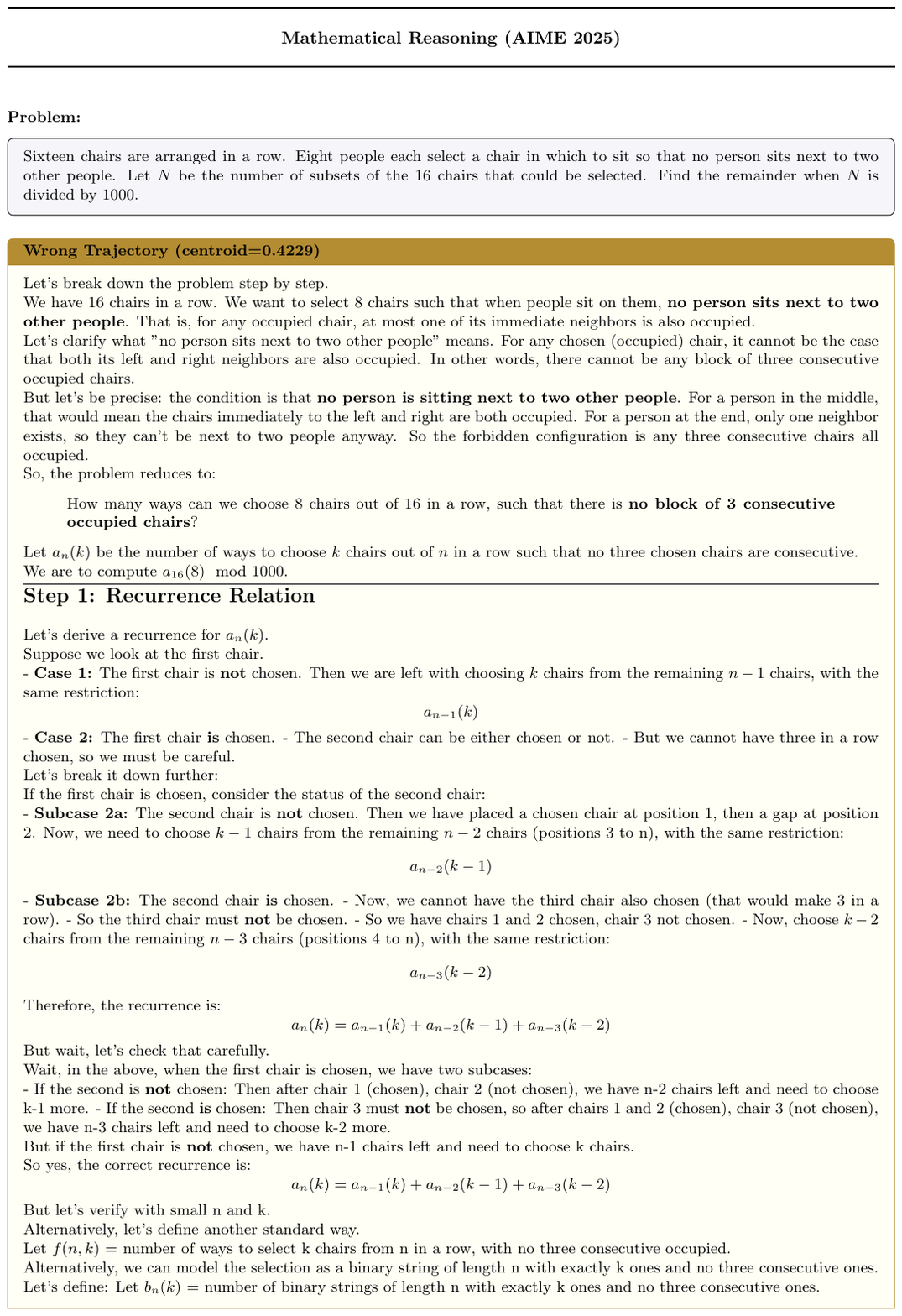}
    \caption{Mathematical Reasoning Task AIME Example. Wrong Trajectory with High Centroid}
    \label{fig:math_wrong}
  \end{figure}

 \begin{figure}[p]
 \ContinuedFloat
    \centering
    \includegraphics[width=\textwidth, page=2]{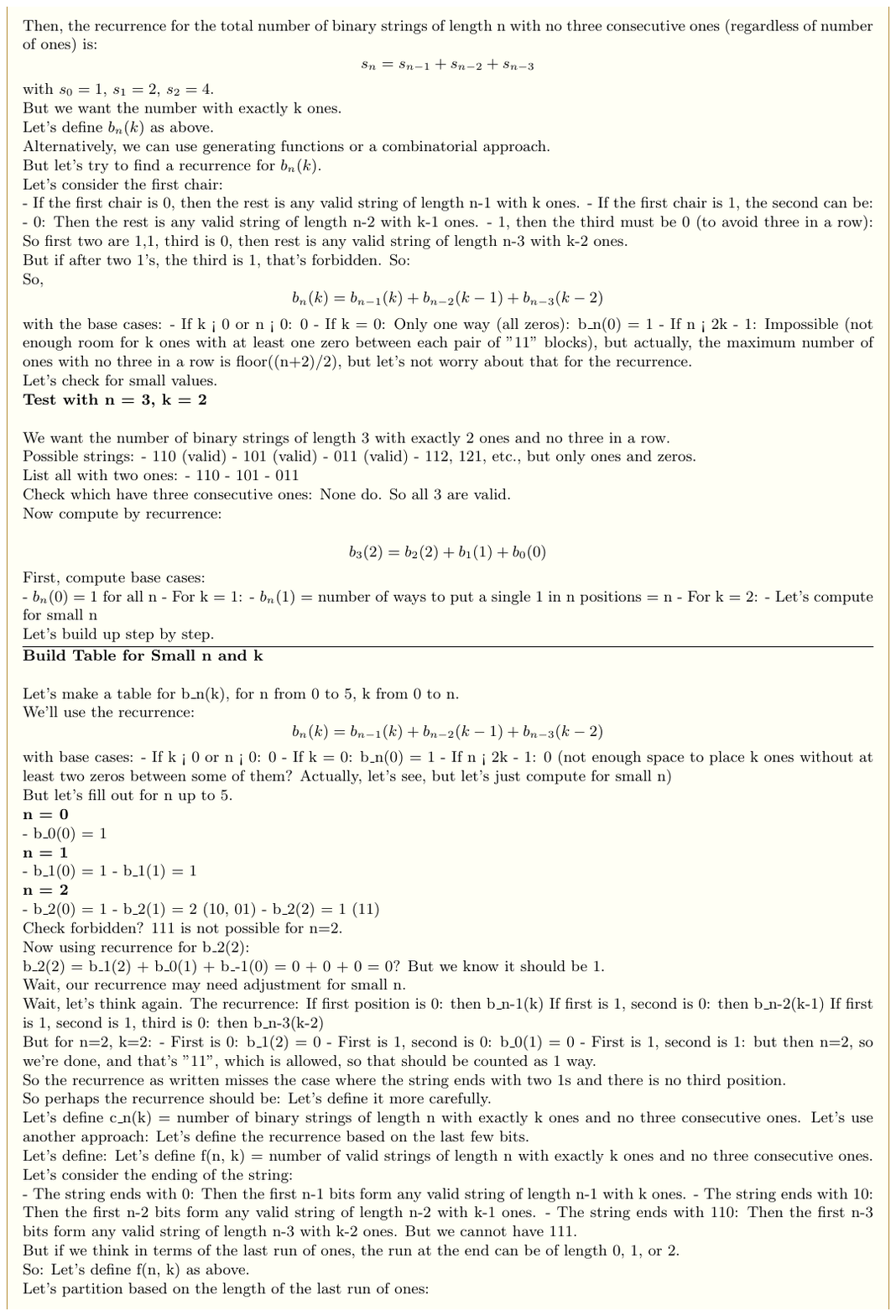}
    \caption{Mathematical Reasoning Task AIME Example. Wrong Trajectory with High Centroid}
  \end{figure}

   \begin{figure}[p]
   \ContinuedFloat
    \centering
    \includegraphics[width=\textwidth, page=3]{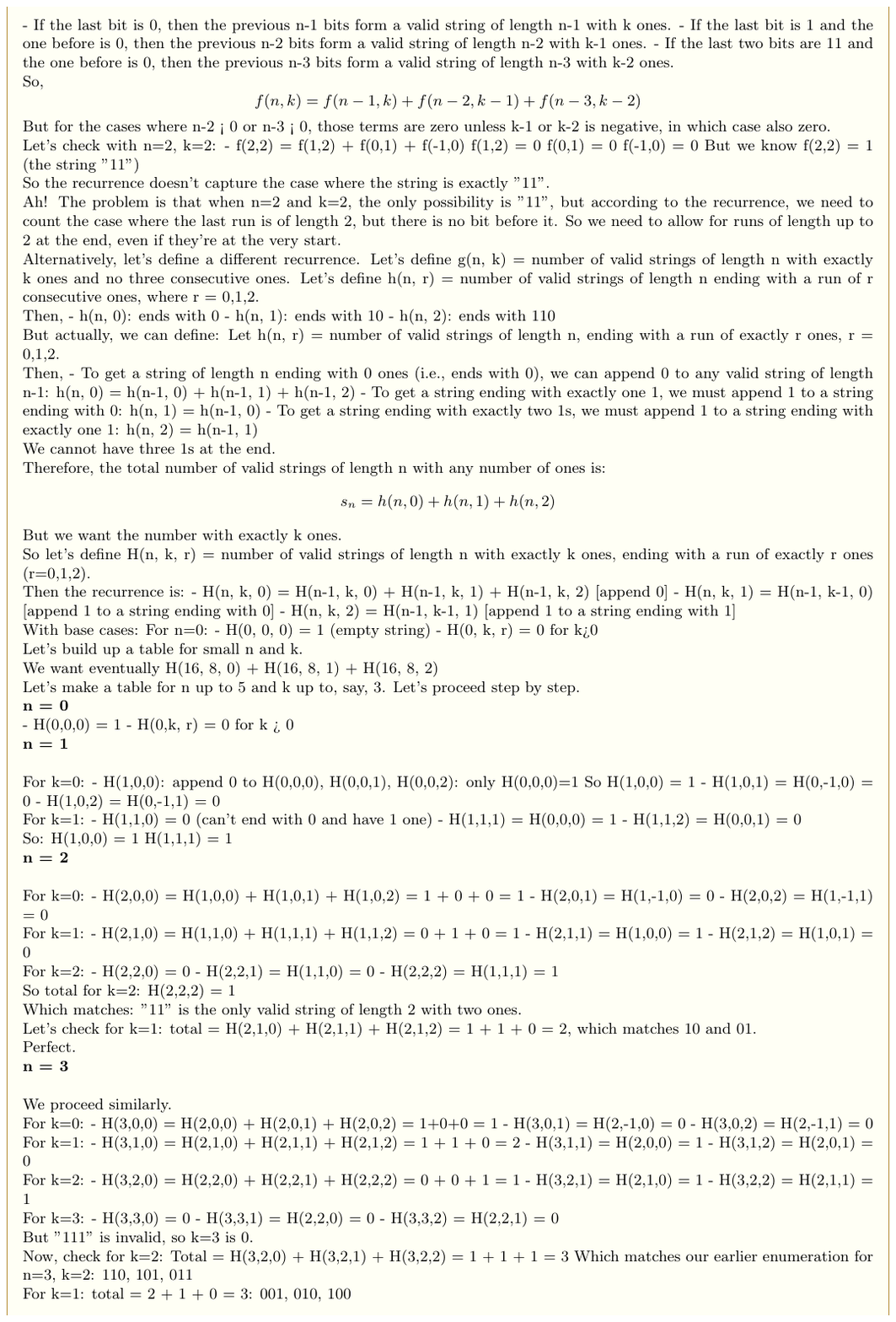}
    \caption{Mathematical Reasoning Task AIME Example. Wrong Trajectory with High Centroid}
  \end{figure}

   \begin{figure}[p]
   \ContinuedFloat
    \centering
    \includegraphics[width=\textwidth, page=4]{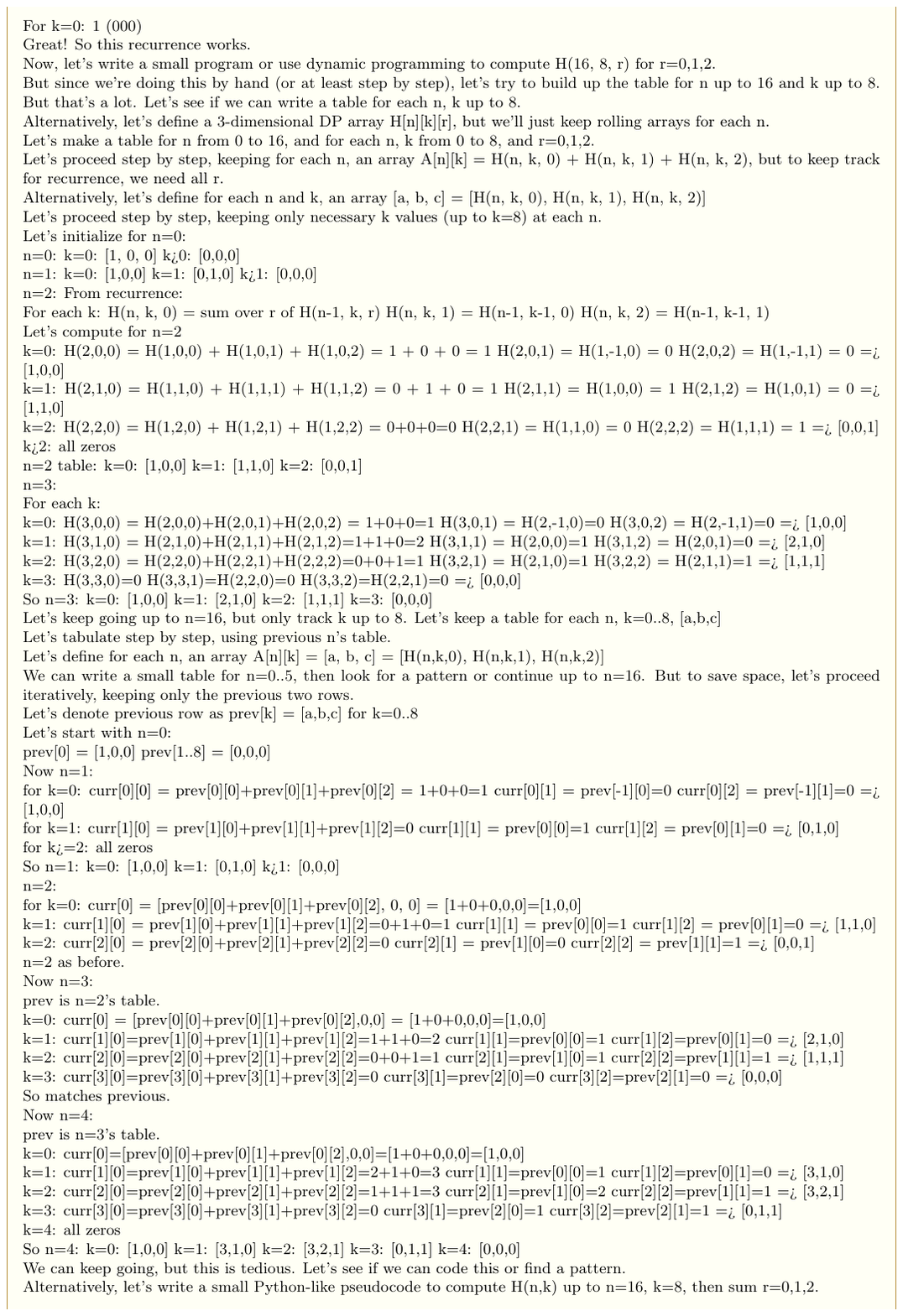}
    \caption{Mathematical Reasoning Task AIME Example. Wrong Trajectory with High Centroid}
  \end{figure}

   \begin{figure}[p]
   \ContinuedFloat
    \centering
    \includegraphics[width=\textwidth, page=5]{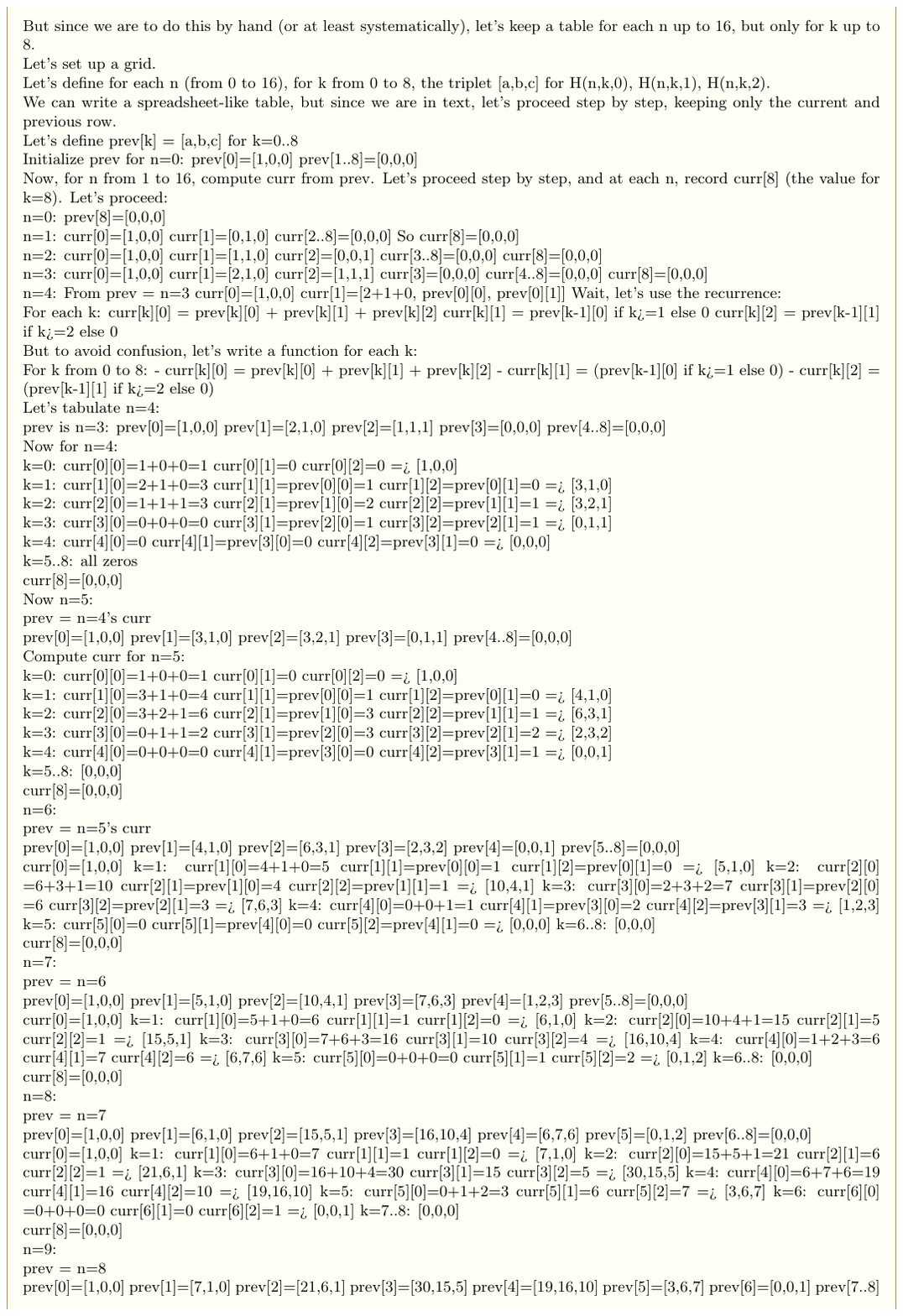}
    \caption{Mathematical Reasoning Task AIME Example. Wrong Trajectory with High Centroid}
  \end{figure}

   \begin{figure}[p]
   \ContinuedFloat
    \centering
    \includegraphics[width=\textwidth, page=6]{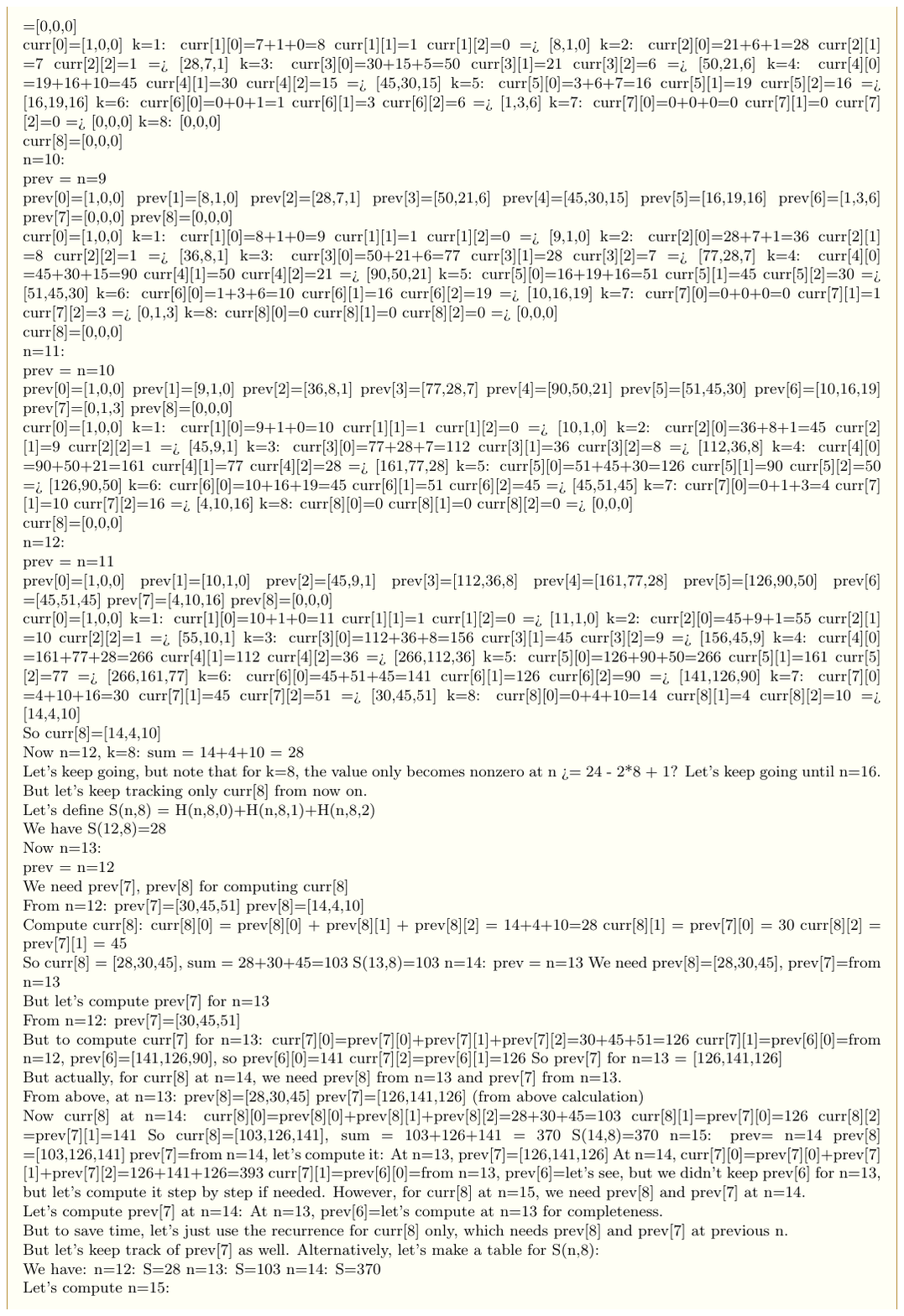}
    \caption{Mathematical Reasoning Task AIME Example. Wrong Trajectory with High Centroid}
  \end{figure}

   \begin{figure}[p]
   \ContinuedFloat
    \centering
    \includegraphics[width=\textwidth, page=7]{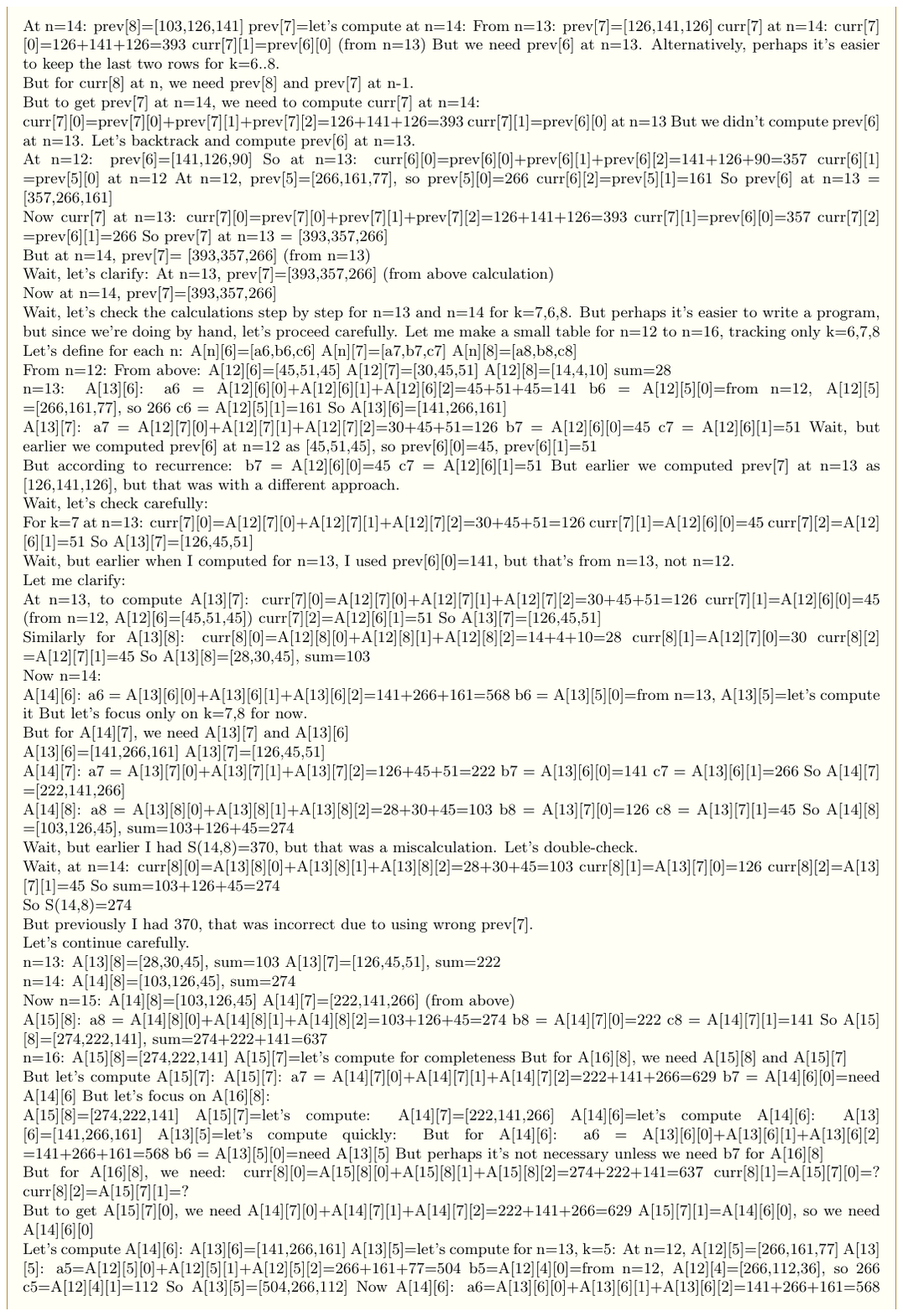}
    \caption{Mathematical Reasoning Task AIME Example. Wrong Trajectory with High Centroid}
  \end{figure}

 \begin{figure}[p]
 \ContinuedFloat
    \centering
    \includegraphics[width=\textwidth, page=8]{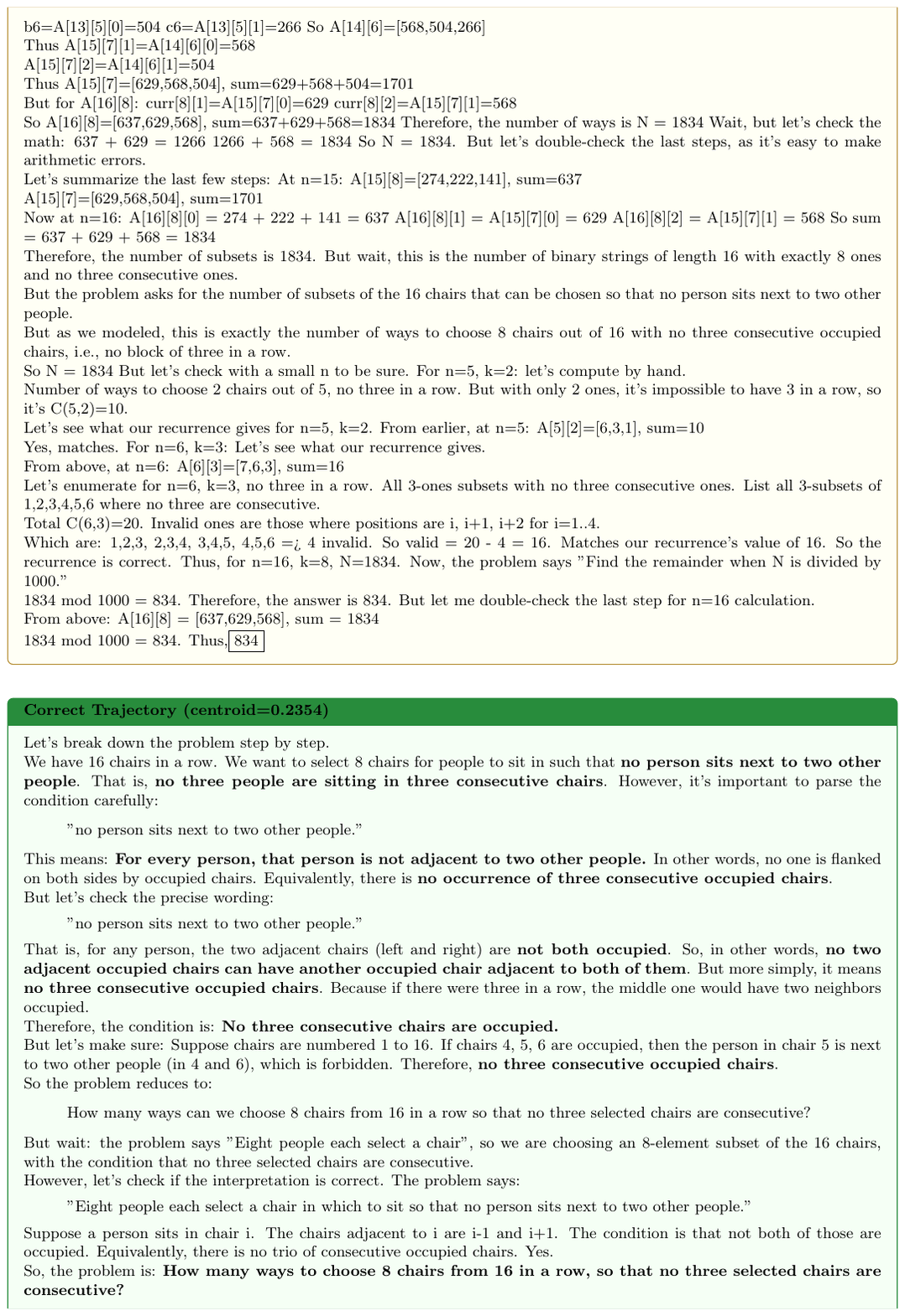}
    \caption{Mathematical Reasoning Task AIME Example. Wrong Trajectory with High Centroid}
  \end{figure}

   \begin{figure}[p]
    \centering
    \includegraphics[width=\textwidth, page=9]{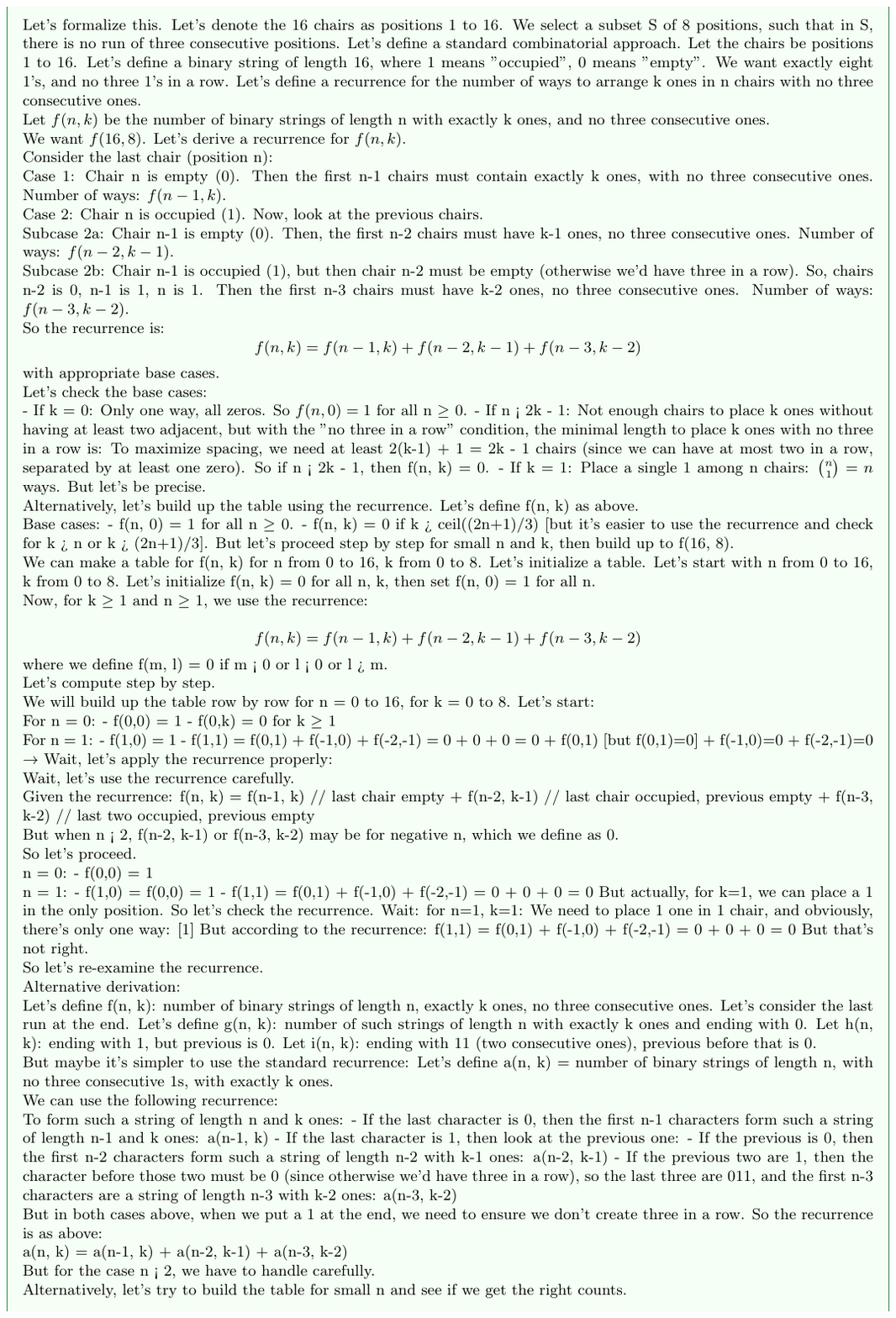}
    \caption{Mathematical Reasoning Task AIME Example. Correct Trajectory with Low Centroid}
    \label{fig:math_correct}

  \end{figure}

     \begin{figure}[p]
     \ContinuedFloat
    \centering
    \includegraphics[width=\textwidth, page=10]{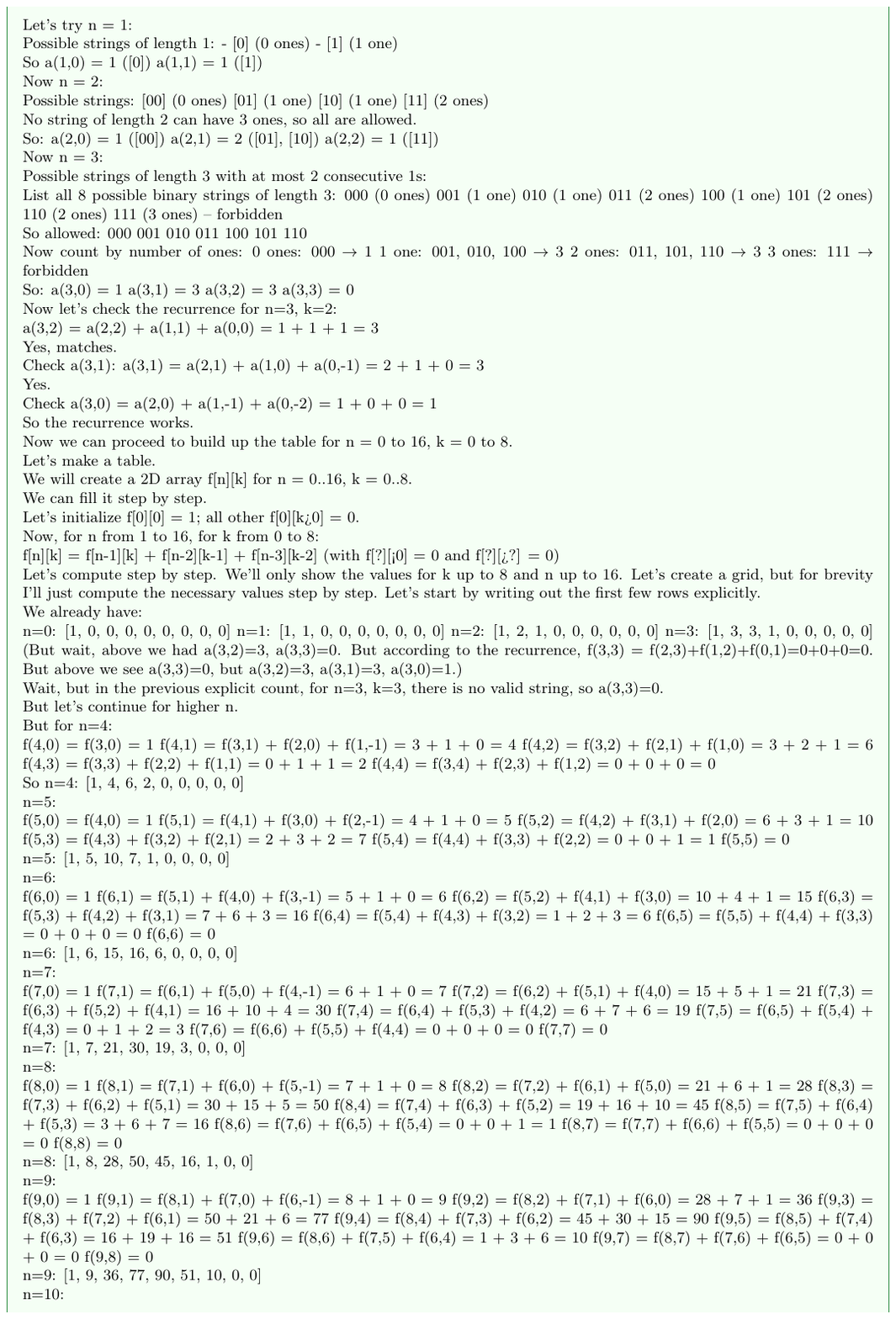}
    \caption{Mathematical Reasoning Task AIME Example. Correct Trajectory with Low Centroid}
  \end{figure}

     \begin{figure}[p]
     \ContinuedFloat
    \centering
    \includegraphics[width=\textwidth, page=11]{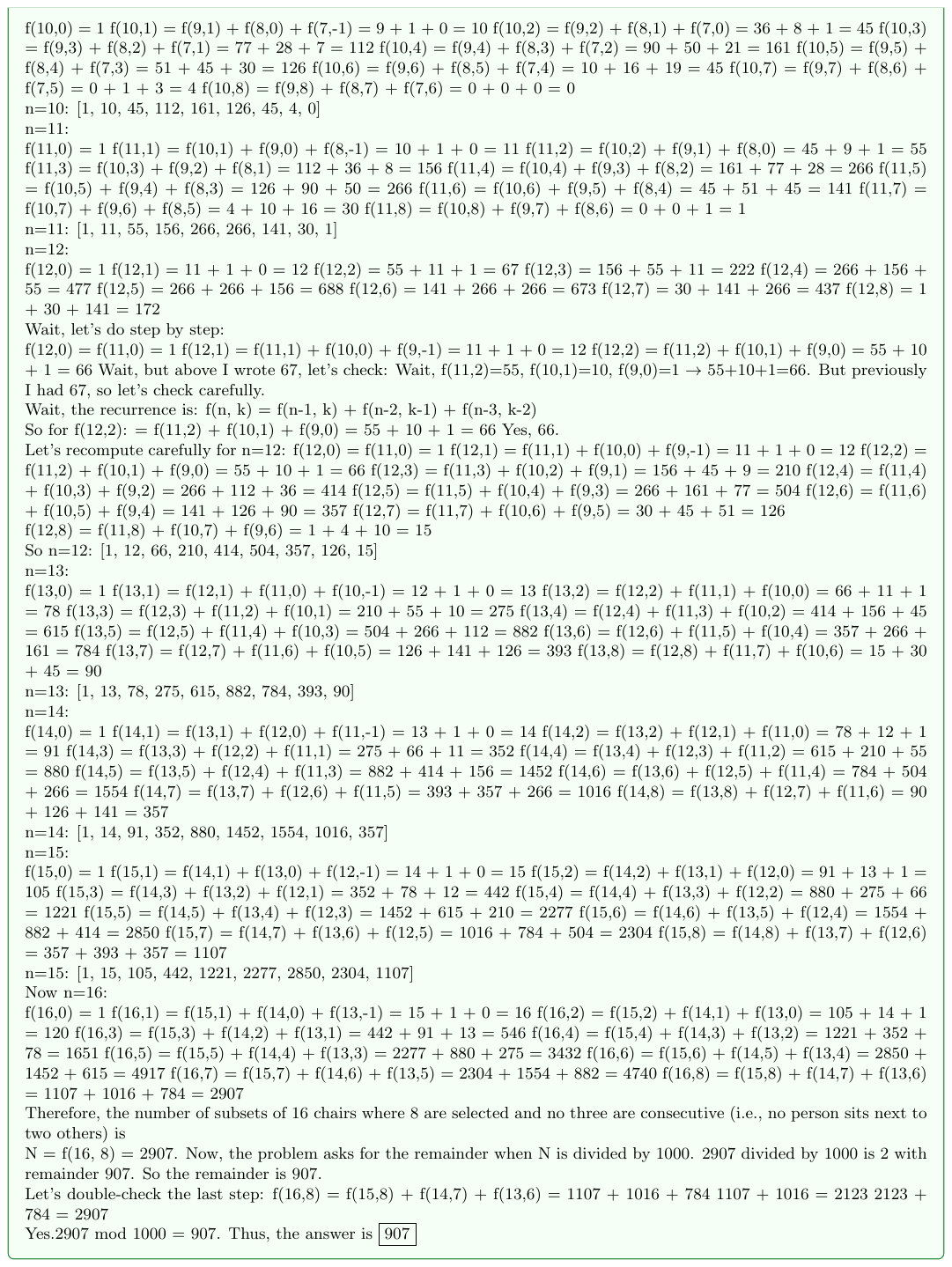}
    \caption{Mathematical Reasoning Task AIME Example. Correct Trajectory with Low Centroid}
  \end{figure}

\end{document}